\documentclass[10pt,twocolumn,letterpaper]{article}

\usepackage[pagenumbers]{cvpr} 

\usepackage{multirow}
\usepackage{booktabs} 
\usepackage{threeparttable}

\definecolor{cvprblue}{rgb}{0.21,0.49,0.74}
\usepackage[pagebackref,breaklinks,colorlinks,allcolors=cvprblue]{hyperref}
\usepackage{microtype}
\usepackage[ruled,vlined]{algorithm2e}
\usepackage{pifont}
\usepackage{graphicx}
\usepackage{adjustbox}
\usepackage{booktabs}   
\usepackage{array}          
\usepackage{arydshln}       
\usepackage{placeins}
\usepackage{hyperref}

\FloatBarrier

\def\paperID{8164} 
\def\confName{CVPR}
\def\confYear{2026}

\title{
    \includegraphics[height=1.2em]{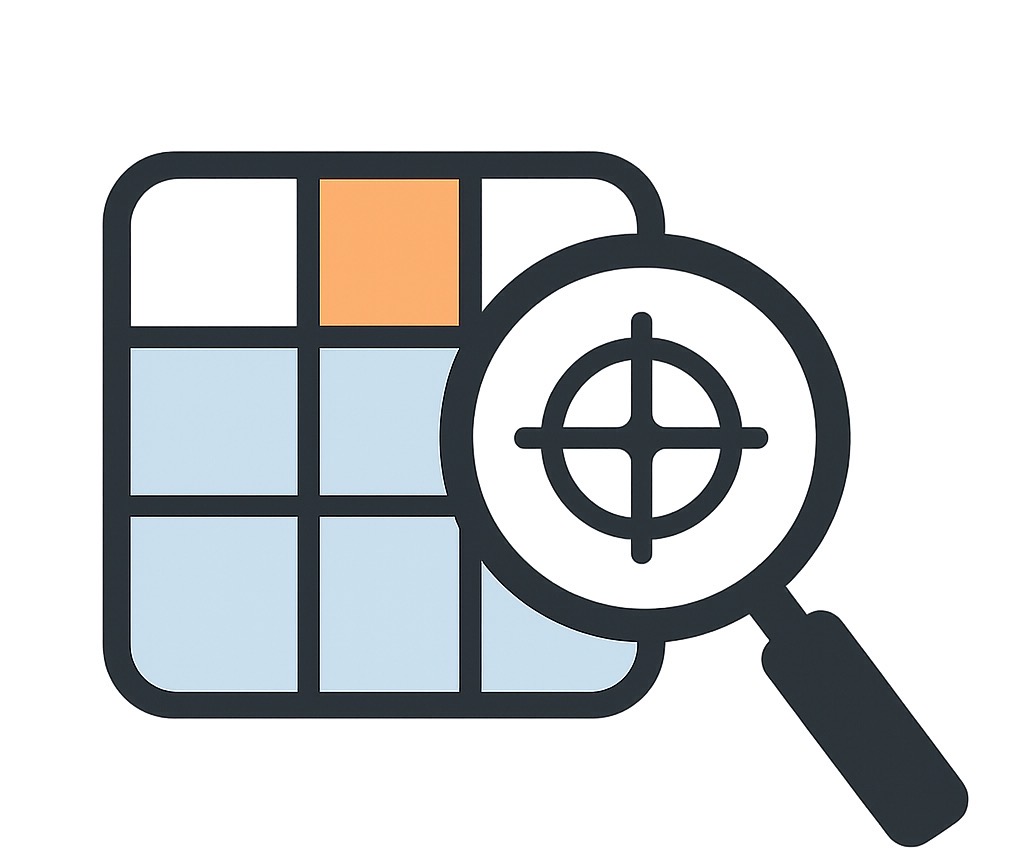} \hspace{0.2em}
    Look Where It Matters: \\
    Training-Free Ultra-HR Remote Sensing VQA via Adaptive Zoom Search
}
\author{
{ \bfseries
Yunqi Zhou\textsuperscript{1*} \quad
Chengjie Jiang\textsuperscript{2*} \quad
Chun Yuan\textsuperscript{2} \quad
Jing Li\textsuperscript{3\textdagger}
}\\
{\small
\textsuperscript{1} Central University of Finance and Economics \quad
\textsuperscript{2} Tsinghua University \quad
\textsuperscript{3} East China Normal University
}
}

\usepackage[table]{xcolor}  

\definecolor{darkred}{RGB}{200, 0, 0}
\definecolor{darkyellow}{RGB}{200, 160, 0}
\definecolor{darkgreen}{RGB}{0, 150, 0}
\definecolor{darkgrey}{RGB}{128, 128, 128}

\begin{document}
\twocolumn[{
\maketitle
\begin{center}
    \includegraphics[width=\textwidth]{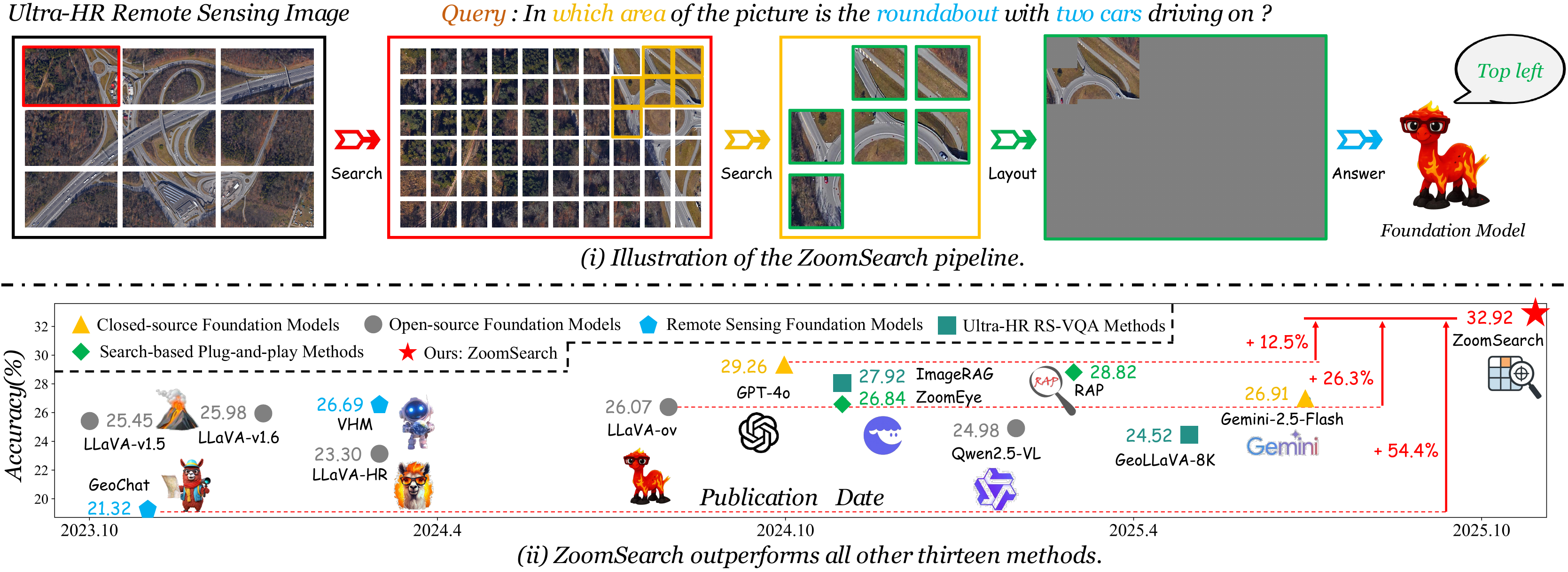}
    \captionof{figure}{
    (i) Illustration of the proposed \textbf{ZoomSearch} pipeline for Ultra-HR RS-VQA. Given an Ultra-HR remote sensing image, ZoomSearch performs a coarse–to-fine zoom-in search, reassembles the selected patches via layout-aware composition, and feeds the results into a foundation model to get the answer.
    (ii) Overall comparison on LRS-VQA. \textbf{ZoomSearch}+LLaVA-ov surpasses all other thirteen methods, achieving an accuracy that is \textbf{12.5\%} higher than the previous best GPT-4o and \textbf{26.3\%} higher than the LLaVA-ov baseline.
    }
    \label{fig:teaser}
\end{center}
}]

\begingroup
\renewcommand{\thefootnote}{\fnsymbol{footnote}} 
\footnotetext[1]{These authors contributed equally to this work.}
\footnotetext[2]{Corresponding author: jingli@geoai.ecnu.edu.cn}
\endgroup

\begin{abstract}
With advances in satellite constellations, sensor technologies, and imaging pipelines, ultra-high-resolution (Ultra-HR) remote sensing imagery is becoming increasingly widespread. However, current remote sensing foundation models are ill-suited to such inputs: full-image encoding exhausts token and memory budgets, while resize-based preprocessing loses fine-grained and answer-critical details. In this context, guiding the model \textbf{look where it matters} before prediction becomes crucial. Therefore, we present \textbf{ZoomSearch}, a training-free, plug-and-play pipeline that decouples `where to look' from `how to answer' for Ultra-HR Remote Sensing Visual Question Answering (RS-VQA). ZoomSearch combines \textbf{Adaptive Multi-Branch Zoom Search}, which performs a hierarchical search over image patches to localize query-relevant regions, with \textbf{Layout-Aware Patch Reassembly}, which reorganizes the selected patches into a compact, layout-faithful canvas. We conduct comprehensive experiments on Ultra-HR RS-VQA benchmarks MME-RealWorld-RS and LRS-VQA, comparing against (i) strong general foundation models, (ii) remote sensing foundation models, (iii) Ultra-HR RS-VQA methods, and (iv) plug-and-play search-based VQA methods. When integrated with LLaVA-ov, \textbf{ZoomSearch} attains state-of-the-art accuracy across diverse tasks, improving the LLaVA-ov baseline by \textbf{26.3\%} on LRS-VQA and \textbf{114.8\%} on MME-RealWorld-RS. Meanwhile, it achieves much higher inference efficiency, outperforming prior search-based methods by \textbf{20$\sim$44\%} in speed. Welcome to visit our \href{https://kiki-zyq.github.io/Zoom-Search/}{project page} for the implementation and additional resources.
\end{abstract}
    
\section{Introduction}
\label{sec:intro}

With advances in low-Earth-orbit satellite constellations~\cite{vleo}, high-performance sensors~\cite{toth2016remotesensing}, and end-to-end imaging pipelines~\cite{wulder}, Ultra-HR imagery has become increasingly prevalent in remote sensing~\cite{dota,isaid,HRVQA,xlrs}. Such data underpin applications in urban monitoring, environmental surveillance, disaster response and so on~\cite{understanding,efficient}. Historically, experts manually interpret these images, which is time-consuming and labor-intensive. Against this backdrop, remote sensing foundation models~\cite{geochat,skysense,skysense-o,skysense++,skysense2,geoground,lhrs,earthgpt,grasp,vhm} have emerged to enable automated analysis and progressed rapidly. They exhibit strong generalization and are now widely applied to RS-VQA, including perception-oriented queries and object-counting tasks. 

However, most of them are trained with fixed low input resolutions (e.g., $336{\times}336$ in VHM~\cite{vhm}; $504{\times}504$ in GeoChat~\cite{geochat}) and are not directly compatible with Ultra-HR imagery (long side $2{\sim}20$K). When applied to Ultra-HR inputs, practitioners typically adopt two handling strategies, both of which are inadequate: (i) \emph{resize-based} preprocessing discards fine structures and small objects that are often answer-critical; and (ii) \emph{tile-and-concatenate} dramatically increases the visual-token budget (e.g., encoding an 8K image at VHM scale requires $\sim$300K visual tokens), resulting in out-of-memory failures and prohibitive inference costs. 

In response to these limitations, recent studies have begun to tackle Ultra-HR RS-VQA along two main directions. The first is \emph{token pruning}~\cite{lrsvqa,geollava8k}, which compresses inputs by removing background or low-information regions. Although effective, such methods typically still rely on downsampling or pooling to maintain computational feasibility of attention, thereby sacrificing detail. The second is \emph{retrieval-augmented generation (RAG)}~\cite{rag,imagerag,rsrag}, which enriches context by retrieving similar patches from labeled databases. However, its effectiveness hinges on retrieval reliability and database–image alignment, both of which can be brittle in remote sensing settings. Despite their distinct limitations, these approaches yield a common and valuable insight: models must \textbf{look where it matters}.

Motivated by this premise, we recast the question of \emph{`where to look'} as an explicit search problem before answer generation. To develop a complete search-based pipeline, we pose two design questions: \textbf{(i) what is an appropriate search unit for Ultra-HR imagery; and (ii) once relevant regions are identified, how should they be reassembled for answer prediction by a foundation model?} To address these questions, we conducted pilot studies on LRS-VQA~\cite{lrsvqa} using LLaVA-ov~\cite{llavaov} as the base model. 

For the \textbf{first question}, we contrasted three search policies: (i) a \emph{fine-tile} policy that evaluates fixed \(224\times224\) tiles directly; (ii) a \emph{coarse-patch} policy that partitions each image into a fixed \(3\times3\) grid and selects among those patches; and (iii) a \emph{hierarchical} policy that recursively splits the currently selected region into a \(3\times3\) grid and continues to zoom in until sufficient evidence is obtained. Owing to substantial variation in object scales across Ultra-HR remote sensing scenes, the hierarchical policy consistently yields superior end-to-end accuracy. To address the \textbf{second question}, we compared three strategies for reassembling the selected regions: (i) \emph{sequential concatenation}, which places patches in a fixed raster order from upper left to lower right; (ii) \emph{relative-layout preserving}, which compacts the canvas while maintaining inter-patch spatial relations; and (iii) \emph{relative \& global-layout preserving}, which additionally preserves each patch’s global orientation with respect to the original image. The third strategy consistently achieves the best results, indicating that retaining spatial cues at both local and global levels is crucial for reliable Ultra-HR RS-VQA.

Guided by the above preliminary studies, we propose \textbf{ZoomSearch}, a training-free, plug-and-play pipeline that separates \emph{`where to look'} from \emph{`how to answer'}. As illustrated in Fig.~\ref{fig:teaser}(i), \textbf{ZoomSearch} first performs a coarse-to-fine zoom-in search: starting from a fixed \(3{\times}3\) grid of the Ultra-HR image, it iteratively zooms into promising areas. At each level, an external scoring model~\cite{visrag, remoteclip, GeoRSCLIP, lrsclip} estimates patch–text relevance, while the foundation-model’s forward confidence provides a model-evidence signal. Candidates are then ranked by these two cues and expanded via an adaptive top-\(k\) strategy to mitigate target dispersion. The selected regions are then reassembled into a compact, model-consumable canvas while preserving spatial cues at both local and global levels. The resulting canvas is finally submitted to a foundation model for answer prediction, without requiring fine-tuning or architectural changes. 

Our main contributions are summarized as follows:

\begin{itemize}
    \item To mitigate the capacity--resolution mismatch in Ultra-HR RS-VQA, we formulate an explicit \textbf{ZoomSearch} paradigm that decouples \emph{`where to look'} from \emph{`how to answer'}, enabling training-free adaptation of existing remote sensing foundation models to $2{\sim}20$K imagery.
    
    \item To achieve this search paradigm in a plug-and-play manner, we design \textbf{Adaptive Multi-Branch Zoom Search} and \textbf{Layout-Aware Patch Reassembly}. The former performs score-guided, coarse-to-fine zoom-in over image patches, while the latter reorganizes the selected regions into a compact canvas that preserves both local topology and global orientation for downstream RS-VQA.

    \item We demonstrate the effectiveness of \textbf{ZoomSearch} on MME-RealWorld-RS and LRS-VQA benchmarks. When plugged into LLaVA-ov, it brings a substantial boost in average accuracy, with \textbf{26.3\%} improvement on LRS-VQA and an impressive \textbf{114.8\%} gain on MME-RealWorld-RS. At the same time, it reduces inference cost by \textbf{20$\sim$44\%} compared with existing search-based pipelines.
\end{itemize}

\section{Related Work}
\label{sec:relatedwork}

\subsection{Ultra-HR Remote Sensing VQA}

\textbf{Token Pruning-based Ultra-HR RS-VQA.}
GeoLLaVA-8K~\cite{geollava8k} proposes \emph{Background Token Pruning} to merge redundant background areas and \emph{Anchored Token Selection} to keep sparse informative objects. However, they both operate on pooled representations to control attention cost, inevitably losing fine details. Luo et al.~\cite{lrsvqa} further introduce text-aware region localization with a coarse-to-fine patch selection scheme that avoids pooling. However, the coarse stage still operates on downsampled images, so pruning decisions are made at low resolution and may miss answer-critical structures, leading to cascading errors. Moreover, most pruning pipelines require additional trainable controllers or backbone modifications, which hinder plug-and-play use.

\textbf{RAG-based Ultra-HR RS-VQA.}
RAG methods~\cite{rag} augment RS-VQA by retrieving semantically related image–text pairs from external labeled Remote Sensing databases and injecting them as additional evidence. ImageRAG~\cite{imagerag} uses RemoteCLIP~\cite{remoteclip} variants to retrieve question-relevant patches and descriptions, which are fed to an RS-MLLM to better handle large, information-dense scenes. RS-RAG~\cite{rsrag} constructs a Remote Sensing World Knowledge corpus and retrieves top-ranked image–text knowledge into knowledge-augmented prompts. However, such approaches depend strongly on the coverage and quality of the external database, which limit the robustness and scalability in practice.

\subsection{Search-Based VQA Methods}

In natural-image VQA, several methods adopt a \emph{locate-then-answer} paradigm~\cite{visualcot,hrrl,dualfocus,DyFo,dc2,vstar,zoomeye}. SEAL~\cite{vstar} performs an LLM-guided visual search that iteratively finds missing targets and stores cropped regions in a visual working memory for answer generation. ZoomEye~\cite{zoomeye} represents an image as a hierarchical tree of zoomed patches and learns a policy to traverse this tree. Retrieval-Augmented Perception (RAP)~\cite{rap} instead treats Ultra-HR VQA as a long-context problem and retrieves only query-relevant crops for answering, trading off context length against coverage.

These search-based approaches show that focusing computation on query-relevant regions is effective for high-resolution natural images. However, remote sensing imagery typically exhibits higher resolutions, cluttered backgrounds, and object-scale variation, calling for a more flexible search strategy. In this context, we propose \textbf{ZoomSearch}, a training-free, plug-and-play framework that operationalizes the principle of \emph{looking where it matters} for Ultra-HR RS-VQA.

\section{Pilot Study}
\label{sec:pilot}
In this section, we conduct a preliminary empirical investigation of the key challenges in designing a search-based framework for Ultra-HR RS-VQA. Specifically, we aim to address two questions: (i) what is an appropriate search unit for Ultra-HR imagery; and (ii) once relevant regions are identified, how should they be reassembled for answer prediction by a foundation model. The insights directly inform the final design of our \textbf{\emph{ZoomSearch}} pipeline presented in the Sec.~\ref{sec:method}.

\subsection{Preliminary}
RS-VQA takes a remote sensing image
$I \in \mathbb{R}^{H \times W \times 3}$ and a natural-language query $q$ as inputs, and requires the model to produce a free-form textual answer $a$. In this work we focus on the Ultra-HR regime, where the image resolution satisfies $2\text{K} \le \max(H, W) \le 20\text{K}$.

Conventional RS-VQA methods directly approximate $f$ with a vision--language backbone $\mathcal{F}$ that ingests the whole image (or a resized/cropped variant),
\begin{equation}
    a = \mathcal{F}(I, q),
\end{equation}
which leads to either excessive visual tokens or loss of fine-grained, answer-critical details.  
Search-based approaches instead adopt a \emph{look-where-it-matters-then-answer} paradigm, first identifying query-relevant regions and then predicting the answer based on these selected visual cues. We decompose the pipeline into:
\begin{equation}
    \mathcal{S}: (I, q) \mapsto \mathcal{P}, \quad
    \mathcal{R}: \mathcal{P} \mapsto \tilde{I}, \quad
    \mathcal{F}: (\tilde{I}, q) \mapsto a,
\end{equation}
where $\mathcal{S}$ is a \emph{search} operator that selects a set of informative regions $\mathcal{P}$ (e.g., image patches) from $I$, $\mathcal{R}$ is a \emph{reassembly} operator that composes the selected regions into a compact canvas $\tilde{I}$, and $\mathcal{F}$ is a foundation model that performs answer prediction based on $\tilde{I}$ and $q$. Our goal in the pilot study is to understand what constitutes an appropriate design for $\mathcal{S}$ and $\mathcal{R}$ in the Ultra-HR regime.

To this end, we conduct experiments on the LRS-VQA~\cite{lrsvqa} dataset. LRS-VQA covers diverse question types, including count, color, category, shape, status, reasoning, rural/urban classification, and target background. Following the original task distribution, we randomly sample $2{,}000$ question--image pairs whose images satisfy $\max(H, W) \ge 2\text{K}$, and use \texttt{llava-onevision-qwen2-0.5b} as the backbone for all subsequent pilot studies on this subset.

\begin{figure}[t]
    \centering
    \includegraphics[width=\columnwidth]{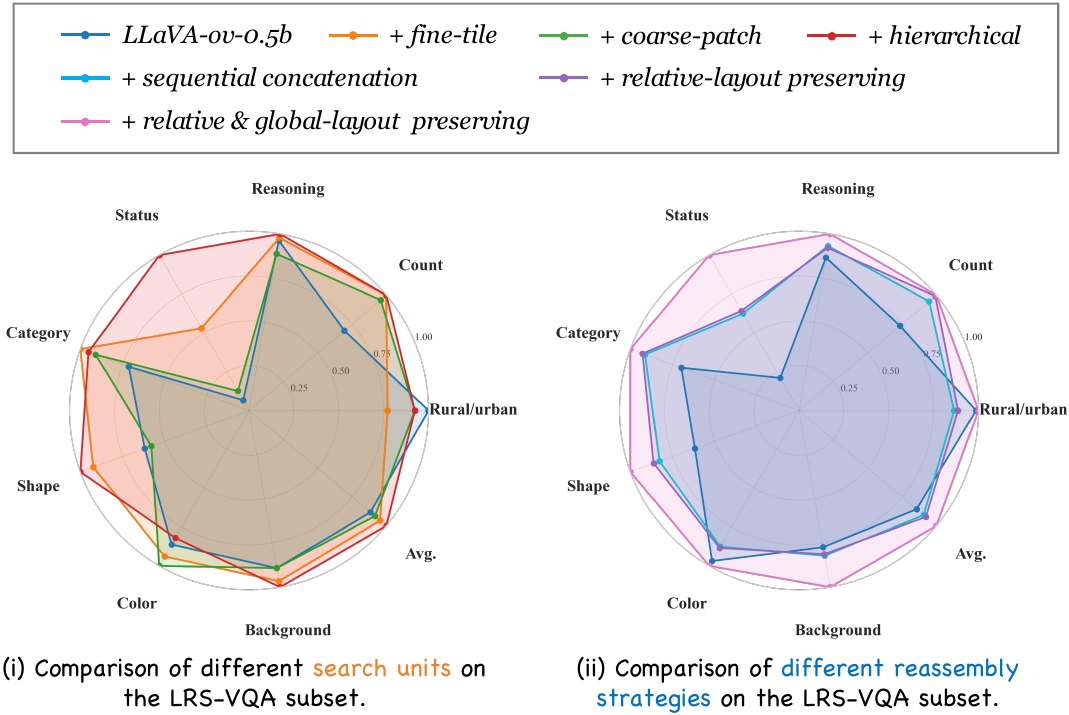}
    \caption{Pilot-study results on the LRS-VQA subset. 
    (i) Among three search units, the \emph{hierarchical} search policy achieves the best performance.
    (ii) Among three reassembly strategies, the \emph{relative \& global-layout preserving} design yields the highest accuracy.
    }
    \label{fig:pilot_lrs}
\end{figure}

\subsection{Exploring Proper Search Units}
\label{subsec:search_units}
To study what constitutes an appropriate search unit for Ultra-HR remote sensing imagery, we compare three search policies: (i) a \emph{fine-tile} policy, (ii) a \emph{coarse-patch} policy and (iii) a \emph{hierarchical} policy. In all cases, the search stage only selects the top-\(1\) patch and feeds it to the foundation model for answer prediction.

We report per-category and average accuracy on the LRS-VQA subset in Fig.~\ref{fig:pilot_lrs}(i).
Overall, the \emph{hierarchical} policy achieves the highest average performance, with particularly large gains on categories that depend on fine object geometry, such as \emph{shape} and \emph{status}. This indicates that progressively zooming from coarse to fine scales is better suited to capturing multi-scale targets in Ultra-HR remote sensing scenes.

\subsection{Exploring Reassembly Strategies}
\label{subsec:reassemly_strategy}

We next investigate how the informative patches should be organized before being fed into the foundation model. In contrast to Sec.~\ref{subsec:search_units}, all methods here use
the hierarchical search policy and select the top-\(k\) patches for each question, which are
then reassembled by one of the following strategies:
(i) \emph{sequential concatenation},
(ii) \emph{relative-layout preserving} and
(iii) \emph{relative \& global-layout preserving}.

Based on Fig.~\ref{fig:pilot_lrs}(ii), the \emph{relative \& global-layout preserving} strategy clearly performs best, achieving the highest accuracy on almost all categories. This suggests that, when multiple informative patches have been located, keeping both their mutual spatial relations and their coarse locations within the original image helps the foundation model better exploit spatial cues after reassembly.
\section{Method}
\label{sec:method}

\begin{figure*}[!t]
    \centering
    \includegraphics[height=0.57\textheight, width=\textwidth]{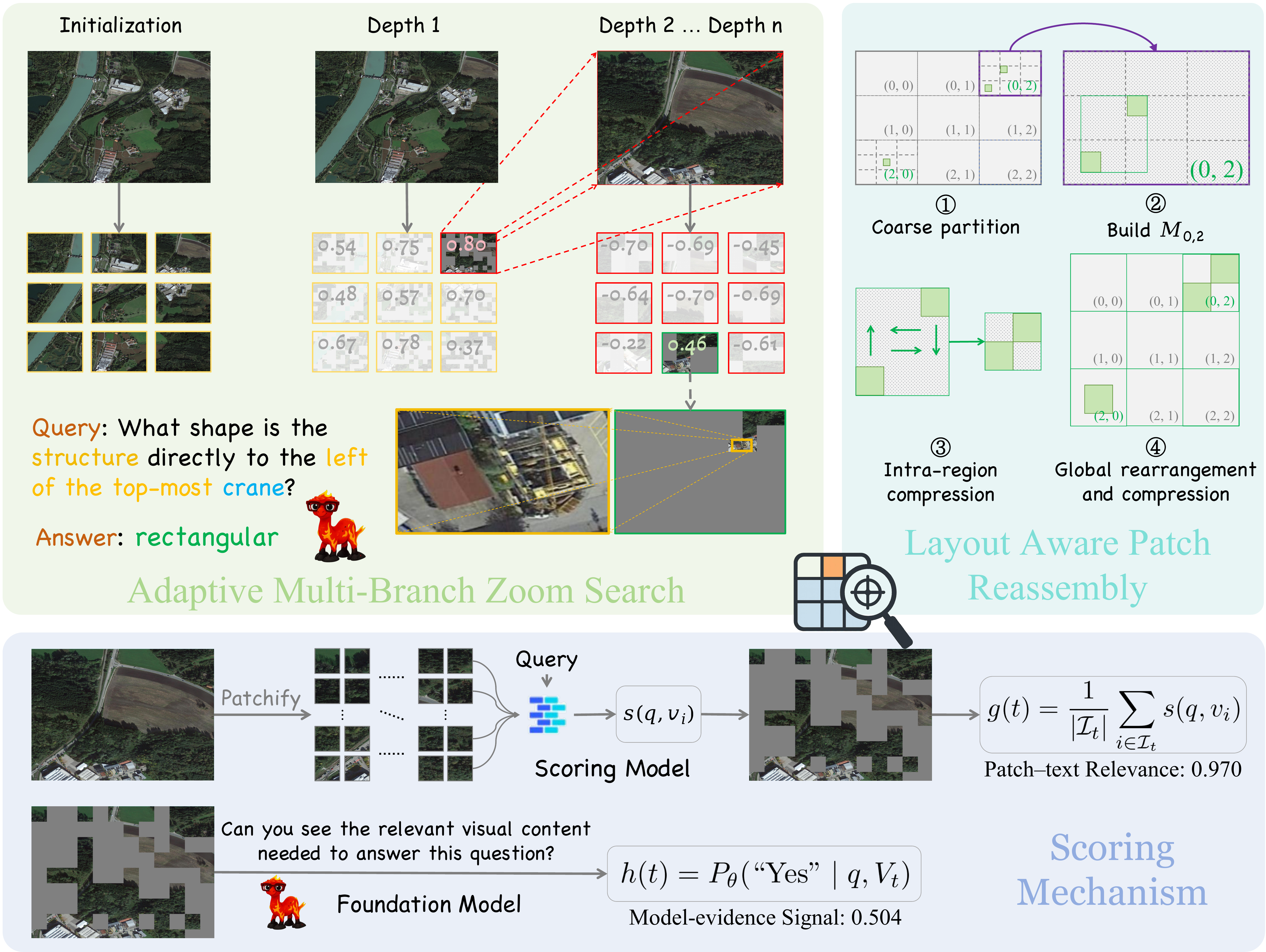}
    \caption{
    Overview of the proposed \textbf{ZoomSearch} pipeline for Ultra-HR RS-VQA.
    The top-left part illustrates \textbf{Adaptive Multi-Branch Zoom Search}, which progressively explores the image and focuses on regions that are closely related to the text query.
    The bottom part shows the \textbf{scoring mechanism}, where each candidate patch is evaluated by a patch--text relevance score from an external scoring model and a model-evidence signal from the foundation model.
    The top-right part depicts \textbf{Layout-Aware Patch Reassembly}, which reorganizes the selected informative patches into a spatially consistent canvas that preserves their relative and global positions.
    }
    \label{fig:zoomsearch_pipeline}
\end{figure*}

\subsection{Method Overview}
Building on the insights from pilot studies in Sec.~\ref{sec:pilot}, we design
\textbf{ZoomSearch} to explicitly decouple \emph{where to look} from \emph{how to answer}
for Ultra-HR RS-VQA. Instead of feeding the entire Ultra-HR image directly
into a foundation model, \textbf{ZoomSearch} first progressively zooms into the scene to
localize question-relevant regions. It then replaces the full-resolution input with a
compact, layout-faithful representation of these regions, which serves as the visual
context for answer prediction. Concretely, we first employ \textbf{Adaptive Multi-Branch Zoom Search} to perform a hierarchical exploration that identifies informative patches at appropriate scales. We then apply \textbf{Layout-Aware Patch Reassembly} to reorganize and compress the selected patches while preserving both local and global spatial relationships. The two components are detailed in the following subsections.

\subsection{Adaptive Multi-Branch Zoom Search}
\label{subsec:zoom_search}

As indicated in Sec.~\ref{subsec:search_units}, hierarchical exploration over image patches is more effective than single-scale search for Ultra-HR RS-VQA. Building on this observation, we design a training-free search procedure, termed \textbf{Adaptive Multi-Branch Zoom Search}, which performs score-guided coarse-to-fine exploration to localize informative regions before answering. The complete algorithm is given in Sec.~\ref{app:algo:search}, while representative search cases are illustrated in the top-left of Fig.~\ref{fig:zoomsearch_pipeline} and in panel (i) of Fig.~\ref{fig:teaser}.



\textbf{Search setup.}
Given an Ultra-HR remote sensing image $I$ and query $q$, we formulate region localization as a tree search. The root is the whole image; a node at depth $d$ is a patch that spawns nine children via a $3{\times}3$ split. The goal is to output the selected-patch set $\mathcal{P}_{\mathrm{sel}}$ that covers the query-relevant regions.

For each candidate patch $t$ at depth $d$, the scoring mechanism in Sec.~\ref{subsec:scoring} produces two quantities: a patch-text relevance score $g(t)$ measuring semantic alignment with $q$, and a model-evidence signal $h(t)$ estimating whether $t$ contains sufficient visual evidence. We fuse $g(t)$ and $h(t)$ into a raw score $f(t;d)$, then apply a sigmoid to all retained patches to obtain $\hat{f}(t;d)$ for ranking.

Equipped with this scoring formulation, we begin the search by dividing $I$ into nine root patches ($d{=}1$). Patches satisfying stopping conditions are added to $\mathcal{P}_{\mathrm{sel}}$; the rest form the initial frontier $\mathcal{F}_1$.

\textbf{Hierarchical expansion with adaptive branching.}
We then iteratively expand the frontier. At each depth $d$, each frontier patch $t$ is
either terminated or subdivided according to three stopping criteria:
(i) the maximum side length of $t$ is no larger than the minimum search unit $u_{\min}$;
(ii) the depth reaches a maximum depth $D_{\max}$; or
(iii) the model-evidence signal exceeds a depth-dependent confidence threshold
$\tau_d = \tau_0 - (d-1)\Delta_\tau$ (with $\tau_0{=}1.0$ and $\Delta_\tau{=}0.1$).
If any criterion is met, $t$ is regarded as sufficiently resolved and appended to
$\mathcal{P}_{\mathrm{sel}}$ without further subdivision.


Otherwise, $t$ is subdivided into a $3{\times}3$ grid of children $\{t_k\}_{k=1}^{9}$ at depth $d+1$. Each child receives a raw score $f(t_k;d+1)$, which is normalized via a sigmoid to produce $\hat{f}(t_k;d+1)$. This normalization enables \emph{adaptive top-$k$ selection} with a fixed threshold $\gamma=0.6$: keep children with $\hat{f}(t_k; d{+}1)>\gamma$ (always retaining the best one if none pass). To bound compute, we cap the kept children by $K_{\max}=6$. The retained nodes form $\mathcal{F}_{d+1}$ and the search proceeds until the frontier empties or a global step budget $S_{\max}$ is met.

In this way, Adaptive Multi-Branch Zoom Search performs a hierarchical
exploration that concentrates computation on high-scoring regions, while the adaptive
top-$k$ branching allows multiple paths to be pursued when the target is spatially
dispersed across different patches. The final selected patch set
$\mathcal{P}_{\mathrm{sel}}$ is passed to the Layout-Aware Patch Reassembly module for spatially faithful composition.

\subsection{Layout-Aware Patch Reassembly}
\label{subsec:layout}
Given the informative patches selected by \textbf{Adaptive Multi-Branch Zoom Search}, our goal is to build a compact canvas that (i) preserves the \emph{relative} layout among patches and (ii) keeps their \emph{global} orientation with respect to the original image. The overall procedure is summarized in Sec.~\ref{app:algo:reassembly} and visualized in the top-right panel of Fig.~\ref{fig:zoomsearch_pipeline}, where steps \ding{172}–\ding{175} correspond to the stages described below.

\textbf{Step \ding{172}: Coarse partition.}
We divide the search grid into a coarse $3{\times}3$ partition $\{\Omega_{i,j}\}_{i,j=0}^2$, and assign each selected informative patch in $\mathcal{P}_{\mathrm{sel}}$ to one coarse region according to its grid indices. For every $(i,j)$, we then build a binary mask $M_{i,j}$ on $\Omega_{i,j}$ that marks which fine-grid cells are selected within that region. This anchors each patch to one of nine quadrants, from top–left to bottom–right, and this coarse spatial layout is preserved in the final canvas.

\textbf{Step \ding{173}–\ding{174}: Intra-region compression.}
Within each coarse region $(i,j)$, we compact its internal layout by removing empty rows
and columns in $M_{i,j}$. Concretely, we compute the sets of non-empty row and column indices
\begin{equation}
    \begin{cases}
    R_{i,j} = \{ r \mid \exists c,\; M_{i,j}(r,c) = 1 \},\\[2pt]
    C_{i,j} = \{ c \mid \exists r,\; M_{i,j}(r,c) = 1 \},
    \end{cases}
\end{equation}
and construct a compact mask $\hat{M}_{i,j}$ by restricting $M_{i,j}$ to $R_{i,j} \times C_{i,j}$. This operation removes only all-zero rows and columns, thereby preserving the order of occupied rows and columns. Consequently, the \emph{relative} arrangement of patches within each coarse region is preserved. Using $\hat{M}_{i,j}$ as an index back to the original image $I$, we crop a compressed content block $I_{i,j}$ with spatial size $(h_{i,j}, w_{i,j})$.

\textbf{Step \ding{175}: Global rearrangement and compression.}
To both preserve \emph{global} positioning and reduce the spatial footprint fed to the
foundation model, we construct a unified cell size that tightly covers all non-empty coarse regions. Over all $(i,j)$ with non-empty $I_{i,j}$, we set
\begin{equation}
    H_c = \max_{i,j} h_{i,j}, \qquad W_c = \max_{i,j} w_{i,j},
\end{equation}
and instantiate a blank $3H_c \times 3W_c$ canvas
$\tilde{I}$ filled with a neutral gray background. For each region $(i,j)$ with non-empty $I_{i,j}$, we resize $I_{i,j}$ to $(H_c, W_c)$ with aspect-ratio–preserving padding and place it into the fixed slot
\begin{equation}
    \tilde{I}[iH_c:(i{+}1)H_c,\; jW_c:(j{+}1)W_c].
\end{equation}
In this way, patches originating from a given coarse region in $I$ are always mapped to the same quadrant in $\tilde{I}$, while their internal spatial arrangement is maintained up to the removal of empty rows/columns. The final canvas $\tilde{I}$ is therefore compact in token budget, preserves the underlying spatial layout, and concentrates the question-relevant content. It is then input to the foundation model for answer prediction.

\begin{table*}[!t]
\centering
\caption{
Quantitative results on the LRS-VQA dataset. \textcolor{darkgreen}{Green} numbers denote the best score, and \textcolor{darkyellow}{yellow} numbers denote the second best. 
The last row (\textit{Improvements}) reports the percentage improvement of \textbf{ZoomSearch (Ours)} over the LLaVA-ov baseline.
}
\label{tab:lrs_dataset}
\resizebox{\textwidth}{!}{%
\begin{tabular}{l c c c c c c c c c c c c}
\toprule
\textbf{Method} & \textbf{Pub.} & \textbf{Max Res.} & \textbf{Rural/Urban} & \textbf{Count} & \textbf{Reasoning} & \textbf{Status} & \textbf{Category} & \textbf{Shape} & \textbf{Color} & \textbf{background} & \textbf{Avg.}\\ 
\midrule
Gemini-2.5-Flash~\cite{gemini} & - & - & 56.49 & \textbf{\textcolor{darkgreen}{18.49}} & 20.40 & \textbf{\textcolor{darkyellow}{18.69}} & 14.81 & 34.17 & 36.08 & 16.11 & 26.91\\
GPT-4o~\cite{gpt4o} & - & - & 55.19 & 16.50 & 21.39 & \textbf{\textcolor{darkgreen}{19.22}} & 16.26 & \textbf{\textcolor{darkgreen}{39.32}} & 47.45 & 18.78 & \textbf{\textcolor{darkyellow}{29.26}}\\
\midrule
LLaVA-v1.5-7b~\cite{llava} & NeurIPS'23 & 336 & 53.04 & 11.84 & 20.50 & 11.80 & 15.40 & 31.98 & 39.87 & 19.18 & 25.45\\
LLaVA-v1.6-7b~\cite{llavanext} & - & 672 & 52.00 & 13.68 & 19.80 & 17.40 & 15.91 & 29.77 & 41.31 & 17.96 & 25.98\\
LLaVA-ov-7b~\cite{llavaov} & TMLR'25 & 384 & 50.08 & 11.68 & 21.80 & 11.20 & 19.15 & 31.98 & 45.49 & 17.14 & 26.07\\
Qwen2.5-VL-7b~\cite{qwen2.5} & - & 1000 & 47.12 & 16.18 & 19.50 & 9.20 & 16.31 & 21.81 & \textbf{\textcolor{darkyellow}{51.76}} & 17.96 & 24.98\\
LLaVA-HR~\cite{luo2024feast} & ICLR'25 & 1536 & 57.11 & 9.67 & 17.60 & 9.10 & 15.20 & 21.02 & 37.91 & 17.96 & 23.30\\
\midrule
GeoChat~\cite{geochat} & CVPR'24 & 504 & 61.42 & 11.76 & 16.70 & 6.50 & 8.00 & 21.47 & 17.39 & 11.84 & 19.38\\
VHM~\cite{vhm} & AAAI'25 & 336 & 56.39 & 12.26 & 18.80 & 13.40 & 17.12 & 31.86 & 46.27 & 12.24 & 26.69\\
\midrule
GeoLLaVA-8K~\cite{geollava8k} & NeurIPS'25 & 8K & 54.68 & 12.83 & 21.32 & 4.41 & 14.81 & 22.41 & 49.52 & 16.18 & 24.52\\
ImageRAG~\cite{imagerag} & GRSM'25 & Dynamic & \textbf{\textcolor{darkyellow}{58.55}} & 13.70 & 21.20 & 10.00 & 21.07 & 33.90 & 45.75 & 19.18 & 27.92\\
\midrule
ZoomEye~\cite{zoomeye} & EMNLP'25 & Dynamic & 48.24 & 14.76 & 22.10 & 10.40 & 23.61 & 28.58 & 45.36 & 21.63 & 26.84\\
RAP~\cite{rap} & ICML'25 & Dynamic & 45.45 & 16.78 & \textbf{\textcolor{darkyellow}{26.10}} & 11.00 & \textbf{\textcolor{darkyellow}{24.32}} & 30.96 & \textbf{\textcolor{darkgreen}{51.90}} & \textbf{\textcolor{darkyellow}{24.08}} & 28.82\\
\midrule
\textbf{ZoomSearch (Ours)} & - & Dynamic & \textbf{\textcolor{darkgreen}{62.53}}  & \textbf{\textcolor{darkyellow}{17.32}} & \textbf{\textcolor{darkgreen}{28.50}} & 15.90 & \textbf{\textcolor{darkgreen}{24.75}} & \textbf{\textcolor{darkyellow}{37.80}} & 50.12 & \textbf{\textcolor{darkgreen}{26.43}}  & \textbf{\textcolor{darkgreen}{32.92}} \\
\textit{Improvements} & - & - & \textbf{\textcolor{darkred}{+ 25\%}}  & \textbf{\textcolor{darkred}{+ 48\%}} & \textbf{\textcolor{darkred}{+ 31\%}} & \textbf{\textcolor{darkred}{+ 42\%}} & \textbf{\textcolor{darkred}{+ 29\%}} & \textbf{\textcolor{darkred}{+ 18\%}} & \textbf{\textcolor{darkred}{+ 10\%}} & \textbf{\textcolor{darkred}{+ 54\%}}  & \textbf{\textcolor{darkred}{+ 26\%}} \\
\bottomrule
\end{tabular}
}
\end{table*}

\subsection{Scoring Mechanism}
\label{subsec:scoring}

For each candidate patch $t$, \textbf{ZoomSearch} computes a scalar score by combining
(i) a \emph{patch--text relevance} term derived from an external scoring model and
(ii) a \emph{model-evidence signal} obtained from the foundation model. We adopt VISRAG~\cite{visrag} as the scoring model, and the overall
pipeline is illustrated in the bottom part of Fig.~\ref{fig:zoomsearch_pipeline}.

\textbf{Patch--text relevance.}
Given a patch $t$ and question $q$, we first decompose $t$ into a set of fixed-size tiles
$\{v_i\}_{i=1}^{n}$, where each $v_i$ has spatial size $u_{\min}\!\times u_{\min}$.
The scoring model $\mathcal{R}$ is a vision–language model trained to estimate
the similarity between a text query and a image unit.
It independently encodes $q$ and each $v_i$ into embedding vectors
$e_q, e_{v_i} \in \mathbb{R}^d$ and computes a cosine similarity score
\begin{equation}
    s(q,v_i)
    = \frac{e_q \cdot e_{v_i}^{\top}}{\|e_q\| \,\|e_{v_i}\|}
\end{equation}

We then sort the tiles by $s(q,v_i)$ and discard the lower 50\% by replacing them with
gray padding, retaining the upper 50\% as potentially relevant evidence.
Remote sensing imagery often contains large homogeneous backgrounds and other redundant
content; keeping only the more relevant half prevents such regions from dominating the
score. The patch--text relevance of $t$ is then defined as the average score over the retained tiles:
\begin{equation}
g(t) = \frac{1}{|\mathcal{I}_t|} \sum_{i \in \mathcal{I}_t} s(q,v_i),
\end{equation}
where $\mathcal{I}_t$ indexes the surviving tiles.

\textbf{Model-evidence signal.}
While $g(t)$ measures the cost already incurred to reach patch $t$, we define a complementary term $h(t)$ to estimate the remaining cost. We first remove rows and columns that are fully padded, obtaining a compact view $V_t$.
We then query $\mathcal{F}$ with $V_t$ and $q$ using a binary prompt
\emph{``Can you see the relevant visual content needed to answer this question?''}
and read out the confidence that the answer is positive.
We define the model-evidence signal
\begin{equation}
h(t) = P_\theta(\text{``Yes''}\mid q,V_t),
\end{equation}
which reflects how likely the backbone believes that $t$ already contains sufficient visual
evidence to answer $q$.

\textbf{Depth-aware fusion.}
To obtain the final score used for search, we blend $g(t)$ and $h(t)$ with a
depth-dependent confidence weight. Following the design in RAP~\cite{rap}, we gradually increase the contribution of the model-evidence term as the search goes deeper, since early in the search the model’s judgment of sufficiency can be unreliable.
For a patch at search depth $d$, we set
\begin{equation}
\omega(d) = (1-b)\Bigl( 1 - \bigl(\tfrac{1}{2}\bigr)^{d-1} \Bigr) + b,
\end{equation}
with bias $b=0.3$. The combined score is
\begin{equation}
f(t;d) = (1-\omega(d))\, g(t) + \omega(d)\, h(t).
\end{equation}
\section{Experiments}
\label{sec:experiments}

\subsection{Experimental Setting}

We evaluate \textbf{ZoomSearch} on two Ultra-HR RS-VQA benchmarks: LRS-VQA~\cite{lrsvqa} and MME-RealWorld-RS~\cite{mmerealworld}. LRS-VQA contains satellite and aerial images whose longest side ranges from $1024$ to $27328$ pixels, and includes eight categories of open-ended questions. MME-RealWorld-RS consists of real-world Ultra-HR remote sensing images with longest side between $689$ and $11500$ pixels, paired with three types of multiple-choice
questions.

Our baselines cover four groups of models.
(i) \textit{Closed-source Foundation Models}: Gemini-2.5-Flash~\cite{gemini} and GPT-4o~\cite{gpt4o}.  
(ii) \textit{Open-source general-purpose Foundation Models}: LLaVA-1.5~\cite{llava}, LLaVA-1.6~\cite{llavanext}, LLaVA-ov~\cite{llavaov}, Qwen2.5-VL~\cite{qwen2.5}, and LLaVA-HR~\cite{llamavid} (specifically trained for higher input resolutions).  
(iii) \textit{Remote-sensing Foundation Models}: GeoChat~\cite{geochat} and VHM~\cite{vhm}.  
(iv) \textit{Ultra-HR RS-VQA methods}: GeoLLaVA-8K~\cite{geollava8k} (token-pruning based) and ImageRAG~\cite{imagerag} (RAG-based).
(v) \textit{Search-based Plug-and-play Methods}: ZoomEye~\cite{zoomeye} and RAP~\cite{rap}. To enable a fair comparison, we build ZoomEye, RAP, and our ZoomSearch on the same base model---\texttt{LLaVA-onevision-qwen2-7b}.

\subsection{Qualitative Results}

\begin{figure}[!b]
    \centering
    \includegraphics[width=\columnwidth]{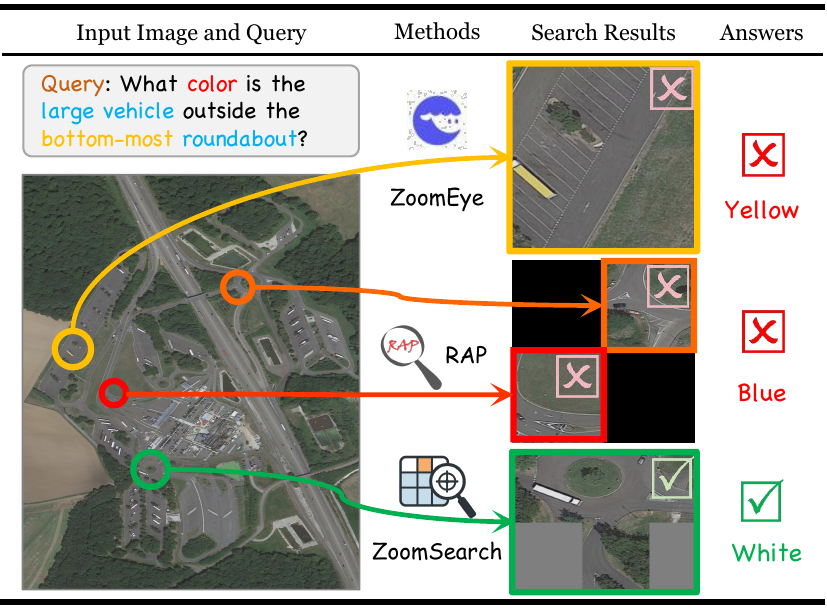} 
    \caption{Qualitative comparison between our method and other search-based methods on an object color recognition task.}
    \label{fig:qualitative_comparison_1}
\end{figure}

\begin{figure}[!b]
    \centering
    \includegraphics[width=\columnwidth]{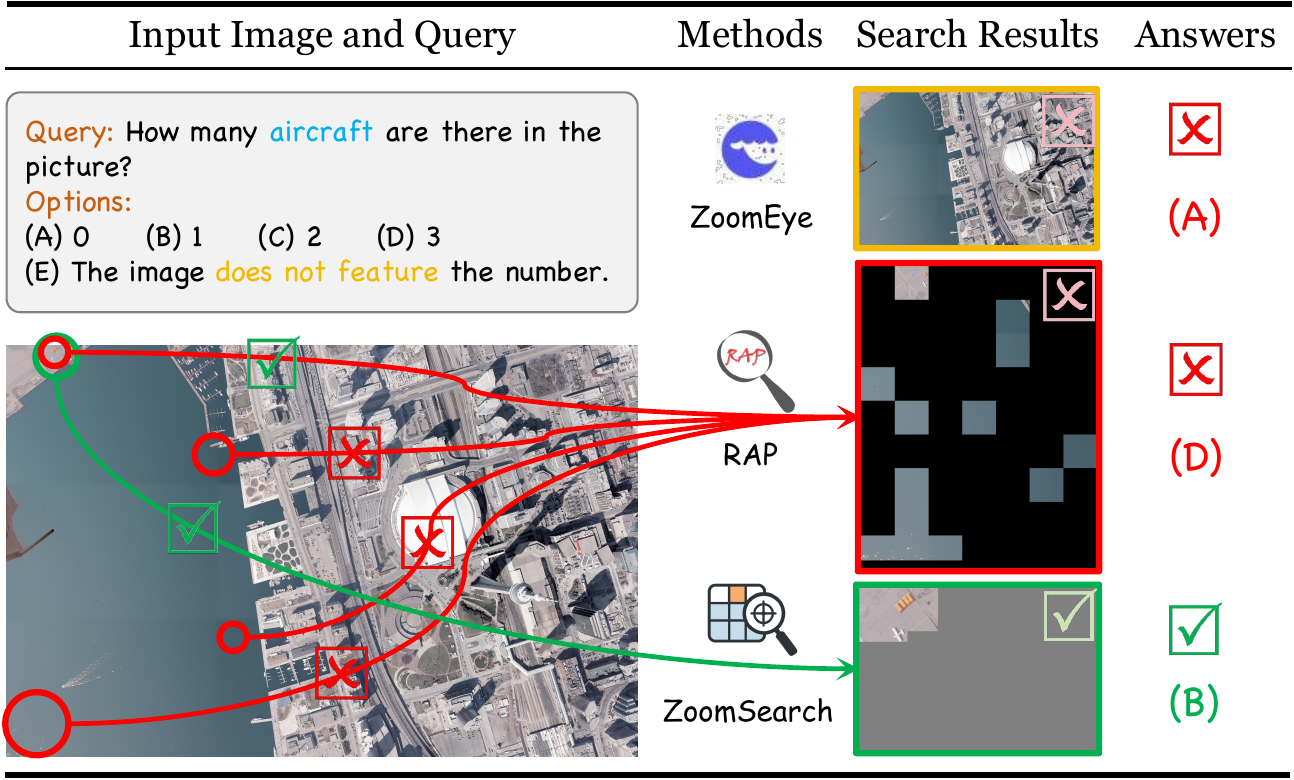} 
    \caption{Qualitative comparison between our method and other search-based methods on an object counting task.}
    \label{fig:qualitative_comparison_2}
\end{figure}

Figs.~\ref{fig:qualitative_comparison_1} and~\ref{fig:qualitative_comparison_2} compare \textbf{ZoomSearch} with state-of-the-art plug-and-play search-based methods on color recognition and counting tasks, respectively. In Fig.~\ref{fig:qualitative_comparison_1}, our method localizes the bottom-most roundabout and the vehicle outside it, and produces the correct color answer, whereas both baselines focus on
irrelevant road segments and fail to answer. 
In the counting example of Fig.~\ref{fig:qualitative_comparison_2}, ZoomEye assigns the highest confidence to the full image itself, causing a failure search. RAP does hit the target region, but simultaneously retrieves many distractors, leading to the final wrong count. In contrast, \textbf{ZoomSearch} precisely zooms into the correct target area, yielding the correct answer. More qualitative examples are provided in panel (i) of Fig.~\ref{fig:teaser}, in Fig.~\ref{fig:zoomsearch_pipeline}, and in Sec.~\ref{app:more_qual}. Overall, these cases highlight two key advantages of our design. First, the hierarchical, adaptive zoom search effectively handles the strong multi-scale nature of remote sensing targets, enabling robust localization of tiny objects in Ultra-HR scenes. Second, the layout-aware reassembly preserves critical spatial cues, which is crucial for queries that depend on fine-grained positional relations.

\subsection{Quantitative Results}

Tables~\ref{tab:lrs_dataset} and~\ref{tab:mme_dataset} report the quantitative results on LRS-VQA and MME-RealWorld-RS.
On LRS-VQA, \textbf{ZoomSearch}+LLaVA-ov reaches SOTA performance on both global tasks such as \emph{Rural/Urban} classification and fine-grained perception tasks such as object \emph{Category}, showing that the hierarchical zoom-in search can capture scene-level context and small-object details simultaneously.
Compared with the LLaVA-ov baseline, ZoomSearch brings substantial gains on almost every category, with relative improvements often exceeding \textbf{20\%}.
These consistent boosts demonstrate that ZoomSearch is highly effective as a plug-and-play enhancement for existing foundation models.

\begin{table}[ht]
\centering
\caption{Quantitative results on MMERealworld-RS dataset.}
\label{tab:mme_dataset}
\resizebox{\columnwidth}{!}{%
\begin{tabular}{l c c c c c}
\toprule
\textbf{Method} & \textbf{Max Res.} & \textbf{Position} & \textbf{Color} & \textbf{Count} & \textbf{Avg.} \\
\midrule
Gemini-2.5-Flash & --  & 55.43 & 50.92 & 29.64 & 45.33  \\
GPT-4o           & --  & 33.52 & 29.83 & 18.90 & 27.42  \\
\midrule
LLaVA-v1.5-7b  & 336  & 21.48 & 22.95 & 16.31 & 20.28 \\
LLaVA-v1.6-7b  & 672  & 26.49 & 24.06 & 20.47 & 23.70 \\
LLaVA-ov-7b   & 384  & 26.81 & 26.14 & 27.57 & 26.83 \\
Qwen2.5-VL-7b & 1000 & 22.12 & 15.54 & 14.93 & 17.55 \\
LLaVA-HR   & 1536 & 35.56 & 44.30 &  7.91 & 29.26 \\
\midrule
GeoChat     & 504 & 25.06 & 23.11 & 15.66 & 21.32 \\
VHM         & 336 & 35.24 & 20.32 & 16.80 & 24.18 \\
\midrule
GeoLLaVA-8K & 8K &   34.90  &   27.92  &   22.27  &   28.41  \\
ImageRAG    & Dynamic & 63.33 & 60.48 & 32.46 & 52.09\\
\midrule
ZoomEye & Dynamic & 43.52 & 60.88 & 30.10 & 44.94 \\
RAP     & Dynamic &  \textbf{\textcolor{darkyellow}{57.62}} & \textbf{\textcolor{darkyellow}{64.53}} & \textbf{\textcolor{darkgreen}{40.25}} & \textbf{\textcolor{darkyellow}{54.20}} \\
\midrule
\textbf{ZoomSearch(Ours)}    & Dynamic &   \textbf{\textcolor{darkgreen}{67.62}}  &  \textbf{\textcolor{darkgreen}{66.14}}  &   \textbf{\textcolor{darkyellow}{39.15}}  &   \textbf{\textcolor{darkgreen}{57.64}}  \\
\textit{Improvements}    & - &   \textbf{\textcolor{darkred}{+ 152\%}}  &  \textbf{\textcolor{darkred}{+ 153\%}}  &   \textbf{\textcolor{darkred}{+ 42\%}}  &   \textbf{\textcolor{darkred}{+ 115\%}}
\\
\bottomrule
\end{tabular}
}
\end{table}

On MME-RealWorld-RS, the gains are especially large on \emph{Position} and \emph{Color}, where accuracy improves by \textbf{152\%} and \textbf{153\%} over LLaVA-ov, respectively.
The former reflects the benefit of our layout-aware reassembly, which preserves both global quadrant placement and relative spatial relationships among patches, while the latter indicates that ZoomSearch can precisely localize the queried object.
The improvement on \emph{Count} (\textbf{+42\%}) is more modest, which we attribute to the limited counting ability of the backbone—correct localization does not always lead to correct answers.

\begin{table}[ht]
\centering
\caption{Efficiency comparison.}
\label{tab:search_methods}
\resizebox{\columnwidth}{!}{%
\begin{tabular}{l cc cc cc}
\toprule
\multirow{2}{*}{\textbf{Method}} & \multicolumn{2}{c}{\textbf{MMERealword-RS}} & \multicolumn{2}{c}{\textbf{LRS-VQA}} & \multicolumn{2}{c}{\textbf{Average}} \\
\cmidrule(lr){2-3} \cmidrule(lr){4-5} \cmidrule(lr){6-7}
& \textbf{Acc.} & \textbf{Time (s)} & \textbf{Acc.} & \textbf{Time (s)} & \textbf{Acc.} & \textbf{Time (s)} \\
\midrule
ZoomEye & 44.94 & 92.72 & 26.84 & 101.86 & 35.89 & 97.29\\
RAP & \textbf{\textcolor{darkyellow}{54.20}} & \textbf{\textcolor{darkyellow}{72.15}} & \textbf{\textcolor{darkyellow}{28.82}} & \textbf{\textcolor{darkyellow}{63.02}} & \textbf{\textcolor{darkyellow}{41.51}} & \textbf{\textcolor{darkyellow}{67.59}}\\
\textbf{ZoomSearch(Ours)} & \textbf{\textcolor{darkgreen}{57.64}} & \textbf{\textcolor{darkgreen}{60.02}} & \textbf{\textcolor{darkgreen}{32.92}} & \textbf{\textcolor{darkgreen}{47.86}} & \textbf{\textcolor{darkgreen}{45.28}} & \textbf{\textcolor{darkgreen}{53.94}}\\
\bottomrule
\end{tabular}
}
\end{table}

Table~\ref{tab:search_methods} further compares inference efficiency with other plug-and-play search-based methods.
On average over the two datasets, \textbf{ZoomSearch} improves accuracy by \textbf{26.2\%} over ZoomEye and \textbf{9.1\%} over RAP, while reducing per-sample latency by \textbf{44.6\%} and \textbf{20.2\%}, respectively.
We attribute these gains to two key design principles: a multi-scale search that rapidly zooms in to query-relevant regions, and an adaptive top-\(k\) retention policy that preserves sufficient branches for comprehensive coverage.

Overall, across two datasets, nine task types, and thirteen strong baselines from four method categories, ZoomSearch+LLaVA-ov consistently attains the highest accuracy by a clear margin: it surpasses the best competing method on LRS-VQA by \textbf{12.5\%} and on MME-RealWorld-RS by \textbf{6.3\%} points in average accuracy, without any retraining or architectural modification of the base model.

\subsection{Ablation Study}
\label{subsec:ablation}

All ablation studies are conducted on the same LRS-VQA subset previously used in the pilot study, with \texttt{LLaVA-onevision-qwen2-7b} as the base model.

\begin{table}[ht]
\centering
\caption{Ablation on Scoring Mechanism.}
\label{tab:ablation_scoring}
\resizebox{\columnwidth}{!}{%
\begin{tabular}{lcccc}
\toprule
\textbf{Method} & \textbf{LRSFAIR} & \textbf{LRSBridge} & \textbf{LRSSTAR} & \textbf{Avg.} \\
\midrule
Baseline (LLaVA-ov-7b)                    & 23.36 & 30.56 & 27.71 & 27.21 \\
Rel.\,+\,Evid.\ (no pruning)               & 26.56 & \textbf{\textcolor{darkyellow}{31.25}} & 33.60 & 30.47 \\
Rel.\,(keep 50\%)+Evid.                  & \textbf{\textcolor{darkgreen}{26.88}} & 30.90 & \textbf{\textcolor{darkyellow}{33.70}} & \textbf{\textcolor{darkyellow}{30.49}} \\
Rel.\,(keep 50\%)+Evid.\,(keep 50\%)     & \textbf{\textcolor{darkyellow}{26.72}} & \textbf{\textcolor{darkgreen}{32.29}} & \textbf{\textcolor{darkgreen}{34.62}} & \textbf{\textcolor{darkgreen}{31.21}} \\
\bottomrule
\end{tabular}
}
\end{table}

\textbf{Scoring mechanism.}
We score each patch with two cues: \emph{patch--text relevance} (Rel.) and \emph{model-evidence} signal (Evid.). 
Table~\ref{tab:ablation_scoring} examines whether we prune irrelevant content before computing each cue (\emph{keep 50\%} means masking the lowest-confidence half for that cue). 
Using both cues without pruning improves the baseline (Avg.\ 30.47 vs.\ 27.21), showing that combining relevance and evidence is beneficial. 
Pruning only Rel.\ yields a slight improvement (Avg.\ 30.49), indicating that background or irrelevant regions indeed interfere with identifying the key object area when computing the relevance score.
By pruning for \emph{both} Rel.\ and Evid., we obtain the best result (Avg.\ \textbf{31.21}). This further indicates that masking non-salient areas when estimating evidence lets the backbone focus on salient content.

\begin{table}[ht]
\centering
\caption{Ablation on branching strategies during the search process.}
\label{tab:topk_comparison}
\tiny
\resizebox{\columnwidth}{!}{%
\begin{tabular}{lcccc}
\toprule
\textbf{Method} & \textbf{LRSFAIR} & \textbf{LRSBridge} & \textbf{LRSSTAR} & \textbf{Avg.} \\
\midrule
Fixed Top-1              & 24.80 & \textbf{\textcolor{darkyellow}{31.94}} & 29.83 & 28.86 \\
Fixed Top-2              & 24.16 & 30.90 & 29.46 & 28.17 \\
Fixed Top-3              & \textbf{\textcolor{darkyellow}{26.08}} & 31.60 & \textbf{\textcolor{darkyellow}{32.87}} & \textbf{\textcolor{darkyellow}{30.18}} \\
Adaptive Top-\(k\)       & \textbf{\textcolor{darkgreen}{26.72}} & \textbf{\textcolor{darkgreen}{32.29}} & \textbf{\textcolor{darkgreen}{34.62}} & \textbf{\textcolor{darkgreen}{31.21}} \\
\bottomrule
\end{tabular}
}
\end{table}

\textbf{Branching strategy.}
Table~\ref{tab:topk_comparison} compares branching strategies during the search process. 
Fixed choices (Top-1/2/3) either miss dispersed targets or waste budget on distractors. In contrast, our \emph{Adaptive Top-$k$} branching expands all children whose normalized score exceeds a fixed threshold, which (i) captures cases where the queried object appears in multiple distant regions, and (ii) recovers cases where queried object is split across several patches by the tiling. This strategy yields the best Avg.\ accuracy \textbf{31.21} (\textcolor{darkred}{+3.18}).

\section{Conclusion}
\label{sec:conclusion}

We introduced \textbf{ZoomSearch}, a training-free, plug-and-play framework for Ultra-HR RS-VQA that (i) conducts \textbf{Adaptive Multi-Branch Zoom Search} to localize query-relevant regions via coarse-to-fine exploration, and (ii) performs \textbf{Layout-Aware Patch Reassembly} to build a compact canvas that preserves both inter-patch relations and global orientation. Extensive experiments demonstrate substantial accuracy gains with faster inference speed than existing methods. These benefits arise from focusing computation \emph{where it matters} through hierarchical zooming, while retaining critical spatial cues under a low visual-token budget. Looking ahead, we plan to further improve both the intelligence and efficiency of the search process for Ultra-HR RS-VQA.

\clearpage
\setcounter{page}{1}           
\renewcommand{\thepage}{S\arabic{page}}  
\renewcommand{\thesection}{S\arabic{section}}
\maketitlesupplementary


\section{Pilot Study: Extended Setup and Analyses}
\label{app:pilot}

\paragraph{Setup.}
We construct a 2{,}000-sample subset from LRS-VQA by first filtering images whose longest side exceeds 2K pixels and then sampling according to the original task distribution. The base model is \texttt{llava-onevision-qwen2-0.5b}. Unless noted, the scoring and search hyper-parameters follow the main paper.

\paragraph{Exploring proper search units.}
Table~\ref{tab:search_units} contrasts three policies. 
The \emph{fine-tile} policy fixes the tile size at $224{\times}224$ and selects the single best tile according to our scoring mechanism. 
The \emph{coarse-patch} policy partitions the image once into a $3{\times}3$ grid and selects the best cell. 
The \emph{hierarchical} policy repeatedly splits the current patch into a $3{\times}3$ grid and zooms in guided by the score, stopping when the evidence threshold is met or the unit size falls to $u_{\min}$.
All three policies outperform the backbone (Avg.\ 24.59\%), confirming that explicitly \emph{looking where it matters} is crucial for Ultra-HR inputs. 
Among them, the hierarchical policy provides the most favorable overall balance: relative to \emph{fine-tile}, it increases average accuracy by \textbf{4.0\%} while reducing latency by \textbf{62.4\%}; relative to \emph{coarse-patch}, it yields a further \textbf{7.6\%} accuracy improvement at a moderate time cost, reflecting a controlled accuracy–efficiency trade-off.

At the category level, the hierarchical strategy delivers clear gains on \emph{Status}, and \emph{Shape}, consistent with its ability to resolve tiny but answer-critical structures.  
An exception is \emph{Color}, for which \emph{coarse-patch} attains the highest score (49.52\%); this suggests that chromatic cues are often better captured at coarser spatial scales, whereas excessive zooming may over-emphasize local textures and introduce noise.

\paragraph{Exploring reassembly strategies.}
We first establish the search protocol: hierarchical exploration with an adaptive Top-$k$ branching strategy using a fixed post-sigmoid threshold of $0.6$. We subsequently compare three strategies for composing the selected patches.
First, \emph{sequential concatenation} encodes patches in raster order and concatenates their tokens. 
Second, \emph{relative-layout preserving} follows RAP~\cite{rap}: it removes only all-zero rows and columns across patches so that pairwise spatial relations are maintained. 
Third, \emph{relative \& global-layout preserving} anchors each coarse region to a fixed quadrant to preserve global orientation, and then applies intra-region compression to preserve the relative layout.

Across task categories, the \emph{relative \& global} strategy attains the highest average accuracy (\textbf{28.38\%}) and yields consistent gains, including \emph{Reasoning} \textbf{26.01\%}, \emph{Status} \textbf{13.48\%}, \emph{Shape} \textbf{37.05\%}, \emph{Color} \textbf{44.59\%}, and \emph{Background} \textbf{20.32\%}. 
These findings suggest that combining quadrant-level anchoring with preservation of intra-region order provides complementary cues, thereby retaining task-relevant information more effectively than alternative layouts.

\section{Algorithmic Details of \textbf{ZoomSearch}}
\label{app:algo}

Due to space constraints in the main paper, this section presents pseudocode for \textbf{ZoomSearch}'s two core algorithms—\textbf{Adaptive Multi-Branch Zoom Search} (Alg.~\ref{alg:amzs}) and \textbf{Layout-Aware Patch Reassembly} (Alg.~\ref{alg:lapr})—together with accompanying explanations.

\subsection{Notation and Inputs}
Given an Ultra-HR remote sensing image \(I\) and a natural-language query \(q\),
\textbf{ZoomSearch} (i) localizes a set of query-relevant patches
\(\mathcal{P}_{\mathrm{sel}}\) and then (ii) composes them into a compact canvas
\(\tilde{I}\) that is fed to a foundation model \(\mathcal{F}\) for answer prediction.
Region scoring during search is provided by an external model \(\mathcal{R}\).

\begin{table}[!t]
\centering
\small
\caption{Hyper-parameter summary for \textbf{ZoomSearch}.}
\label{tab:hparams}
\begin{adjustbox}{max width=\columnwidth}
\begin{tabular}{lll}
\toprule
\textbf{Symbol} & \textbf{Role} & \textbf{Value / Rule} \\
\midrule
$u_{\min}$ & Minimum search unit (leaf size) & See Tab.~\ref{tab:umin_rule} \\
$D_{\max}$ & Maximum depth & 50 (safety cap) \\
$S_{\max}$ & Step budget (node expansions) & Per-experiment cap \\
$K_{\max}$ & Max children kept per expansion & 6 \\
$\gamma$ & Keep threshold for normalized score & 0.6 \\
$\tau_d$ & Evidence stop threshold & $\tau_0-(d-1)\Delta\tau$ \\
$\tau_0,\ \Delta\tau$ & Init \& decay for $\tau_d$ & 1.0,\ 0.1 \\
$b$ & Depth-bias in score fusion & 0.3 \\
\bottomrule
\end{tabular}
\end{adjustbox}
\end{table}

\begin{table}[!t]
\centering
\footnotesize
\caption{Rule for choosing the minimum search unit $u_{\min}$.}
\label{tab:umin_rule}
\begin{adjustbox}{max width=\columnwidth}
\begin{tabular}{lll}
\toprule
\textbf{Backbone} & \textbf{Condition on $\max(H,W)$} & \(\boldsymbol{u_{\min}}\) \\
\midrule
\texttt{LLaVA-OneVision}~\cite{llavaov} & $\max(H,W)\ \ge\ 4000$ & 448 \\
\texttt{LLaVA-OneVision} & $\max(H,W)\ <\ 4000$ & 112 \\
Non-\texttt{LLaVA-OneVision}   & $\max(H,W)\ >\ 5096$ & 336 \\
Non-\texttt{LLaVA-OneVision}   & otherwise             & 224 \\
\bottomrule
\end{tabular}
\end{adjustbox}
\end{table}

\begin{table*}[!t]
\centering
\caption{Comparison of different search units on the LRS-VQA subset.}
\label{tab:search_units}
\resizebox{\textwidth}{!}{%
\begin{tabular}{lcccccccccc}
\toprule
\textbf{Method} &
\textbf{Rural/urban} &
\textbf{Count} &
\textbf{Reasoning} &
\textbf{Status} &
\textbf{Category} &
\textbf{Shape} &
\textbf{Color} &
\textbf{Background} &
\textbf{Time} &
\textbf{Avg.} \\
\midrule
LLaVA-ov-0.5b   & 58.77\% & 12.84\% & 22.79\% & 4.41\% & 14.81\% & 23.65\% & 43.27\% & 16.18\% & 2.94s & 24.59\% \\
\midrule
+ fine-tile       & 47.34\% & \textbf{\textcolor{darkyellow}{16.72\%}} & \textbf{\textcolor{darkyellow}{23.11\%}} & \textbf{\textcolor{darkyellow}{7.31\%}} & \textbf{\textcolor{darkgreen}{19.14\%}} & \textbf{\textcolor{darkyellow}{33.33\%}} & \textbf{\textcolor{darkyellow}{46.77\%}} & \textbf{\textcolor{darkyellow}{17.19\%}} & 175.89s & \textbf{\textcolor{darkyellow}{26.24\%}}      \\
+ coarse-patch    & \textbf{\textcolor{darkgreen}{54.68\%}} & 16.21\% & 21.32\% & 4.78\% & 17.78\% & 22.41\% & \textbf{\textcolor{darkgreen}{49.52\%}} & 16.18\% & \textbf{\textcolor{darkgreen}{7.61s}} & 25.36\% \\
+ hierarchical    & \textbf{\textcolor{darkgreen}{54.68\%}} & \textbf{\textcolor{darkgreen}{16.74\%}} & \textbf{\textcolor{darkgreen}{23.53\%}} & \textbf{\textcolor{darkgreen}{10.25\%}} & \textbf{\textcolor{darkyellow}{18.41\%}} & \textbf{\textcolor{darkgreen}{35.76\%}} & 41.35\% & \textbf{\textcolor{darkgreen}{17.65\%}} & \textbf{\textcolor{darkyellow}{66.15s}} & \textbf{\textcolor{darkgreen}{27.30\%}}       \\
\bottomrule
\end{tabular}%
}
\end{table*}

\begin{table*}[!t]
\centering
\caption{Comparison of different reassembly strategies on the LRS-VQA subset.}
\label{tab:reassembly_strategies}
\resizebox{\textwidth}{!}{%
\begin{tabular}{lccccccccc}
\toprule
\textbf{Method} &
\textbf{Rural/urban} &
\textbf{Count} &
\textbf{Reasoning} &
\textbf{Status} &
\textbf{Category} &
\textbf{Shape} &
\textbf{Color} &
\textbf{Background} &
\textbf{Avg.} \\
\midrule
LLaVA-ov-0.5b   & \textbf{\textcolor{darkyellow}{58.77\%}} & 12.84\% & 22.79\% & 4.41\% & 14.81\% & 23.65\% & \textbf{\textcolor{darkyellow}{43.27\%}} & 16.18\% & 24.59\% \\
\midrule
+ sequential concatenation                & 51.77\% & 15.96\% & \textbf{\textcolor{darkyellow}{24.38\%}} & 9.17\% & 18.74\% & 31.01\% & 39.34\% & \textbf{\textcolor{darkyellow}{17.06\%}} & 25.93\%       \\
+ relative-layout preserving              & 53.04\% & \textbf{\textcolor{darkyellow}{16.63\%}} & 24.12\% & \textbf{\textcolor{darkyellow}{9.37\%}} & \textbf{\textcolor{darkyellow}{19.01\%}} & \textbf{\textcolor{darkyellow}{32.22\%}} & 39.68\% & 16.90\% & \textbf{\textcolor{darkyellow}{26.37\%}} \\
+ relative \& global-layout preserving    & \textbf{\textcolor{darkgreen}{59.53\%}} & \textbf{\textcolor{darkgreen}{16.74\%}} & \textbf{\textcolor{darkgreen}{26.01\%}} & \textbf{\textcolor{darkgreen}{13.48\%}} & \textbf{\textcolor{darkgreen}{20.33\%}} & \textbf{\textcolor{darkgreen}{37.05\%}} & \textbf{\textcolor{darkgreen}{44.59\%}} & \textbf{\textcolor{darkgreen}{20.32\%}} & \textbf{\textcolor{darkgreen}{28.38\%}}      \\
\bottomrule
\end{tabular}%
}
\end{table*}

\subsection{Hyper-parameters}

This subsection summarizes all hyper-parameters used by \textbf{ZoomSearch}. 
Table~\ref{tab:hparams} lists the symbols, roles, and default values, while 
Table~\ref{tab:umin_rule} details the adaptive rule for choosing the minimum search unit $u_{\min}$ by backbone and image size.

\subsection{Patch Scoring}
\label{app:algo:scoring}
For a patch $p$ at depth $d$, we compute two cues: a patch--text relevance score $g(p)$ and a model–evidence signal score $h(p)$. We then apply a depth-aware fusion that gradually increases the contribution of $h$ as the search goes deeper:
\begin{equation}
  \omega(d) = (1-b)\Bigl(1-\bigl(\tfrac{1}{2}\bigr)^{d-1}\Bigr) + b,
\end{equation}
and obtain the fused score
\begin{equation}
  f(p;d) = \bigl(1-\omega(d)\bigr)\, g(p) + \omega(d)\, h(p).
\end{equation}
A single sigmoid is applied to $f(p;d)$ before comparison, yielding the normalized score
$\hat{s}(p)=\sigma\!\bigl(f(p;d)\bigr)$, which is used for ranking and for the fixed-threshold Adaptive Top-$k$ selection in Alg.~\ref{alg:amzs}.

\subsection{Adaptive Multi-Branch Zoom Search}
\label{app:algo:search}

\noindent\textbf{Overview and link to tree search.}
Our procedure follows the spirit of A*: we keep an open list (priority queue) $\mathcal{O}$, rank candidates by a fused score $\hat{s}$ (analogous to an $f$-score), and iteratively expand the best frontier while committing finished nodes to $\mathcal{P}_{\mathrm{sel}}$. The score combines two cues, $g$ and $h$ (Sec.~\ref{app:algo:scoring}), and a sigmoid normalization makes a fixed threshold behave adaptively within each set of sibling patches.

\noindent\textbf{Initialization.}
We split $I$ into a $3{\times}3$ grid to obtain nine root patches at depth $1$ and compute $\hat{s}$ and $h$ for each. If a root already satisfies any stopping criterion (size $\le u_{\min}$ or $h\!\ge\!\tau_1$), it is appended to $\mathcal{P}_{\mathrm{sel}}$; otherwise, it is inserted into the priority queue $\mathcal{O}$ keyed by $\hat{s}$.

\noindent\textbf{Stopping criteria.}
At each iteration, we pop the highest-$\hat{s}$ node $(p,d)$ from $\mathcal{O}$ and halt its expansion under \emph{any} of the following conditions: 
(i) depth cap — $d \ge D_{\max}$; 
(ii) leaf-size reached — $\max(\mathrm{height}(p),\mathrm{width}(p)) \le u_{\min}$; 
(iii) evidence-sufficient — $h(p)\!\ge\!\tau_d$. 
Nodes stopped by any rule are \emph{committed} to the selected set $\mathcal{P}_{\mathrm{sel}}$.

\noindent\textbf{Expansion with adaptive branching.}
We expand the frontier patch $p$ by splitting it into a $3{\times}3$ set of children $\{p_k\}_{k=1}^9$ at depth $d{+}1$, and score each child to obtain $(\hat{s}_k, h_k)$. 
Children with $\hat{s}_k \ge \gamma$ are retained; if none meet the threshold, the top-$1$ child is kept to guarantee progress. 
The retained set is truncated to at most $K_{\max}$ nodes and each is enqueued into $\mathcal{O}$ with priority $\hat{s}_k$. 
The loop terminates when $\mathcal{O}$ is empty or the global step budget reaches $S_{\max}$, at which point the algorithm returns $\mathcal{P}_{\mathrm{sel}}$ (Alg.~\ref{alg:amzs}).

\begin{algorithm*}[ht]
\caption{Adaptive Multi-Branch Zoom Search}
\label{alg:amzs}
\small
\DontPrintSemicolon

\KwIn{Ultra-HR image $I$, question $q$, scoring model $\mathcal{R}$, foundation model $\mathcal{F}$,
      minimum search unit $u_{\min}$, maximum depth $D_{\max}$, maximum steps $S_{\max}$,
      branch cap $K_{\max}$, score threshold $\gamma$, evidence schedule $\{\tau_d\}$, bias $b$}
\KwOut{selected informative patch set $\mathcal{P}_{\mathrm{sel}}$}

\BlankLine
\textbf{Function} GetWeight$(d,b)$\;
\Indp
\Return $(1-b)\!\cdot\!\bigl(1-(\tfrac{1}{2})^{\,d-1}\bigr)+b$\;
\Indm

\BlankLine
\textbf{Function} ScorePatch$(p,q,d)$\;
\Indp
$g \leftarrow \textsc{TextRelevance}(p,q;\mathcal{R},u_{\min})$\;
$p' \leftarrow \textsc{PrunePatch}(p,u_{\min})$\;
$h \leftarrow \textsc{ModelEvidence}(p',q;\mathcal{F})$\;
$\omega \leftarrow$ GetWeight$(d,b)$,\quad $f \leftarrow (1-\omega)\,g+\omega\,h$\;
$\hat{s} \leftarrow \sigma(f)$;\ \Return $(\hat{s},h)$\;
\Indm

\BlankLine
\textbf{/* initialization */}\;
Divide $I$ into a $3{\times}3$ grid to get $\{p_j\}_{j=1}^9$;\quad
$\mathcal{O}\!\leftarrow\!\emptyset$, $\mathcal{P}_{\mathrm{sel}}\!\leftarrow\!\emptyset$, $S\!\leftarrow\!0$\;
\For{$j=1$ \KwTo $9$}{
  $(\hat{s},h)\leftarrow$ ScorePatch$(p_j,q,1)$\;
  \If{$\max(\mathrm{height}(p_j),\mathrm{width}(p_j)) \le u_{\min}$ \textbf{or} $h \ge \tau_1$}{
    $\mathcal{P}_{\mathrm{sel}} \leftarrow \mathcal{P}_{\mathrm{sel}} \cup \{p_j\}$}
  \Else{ insert node $(p_j,1,\hat{s},h)$ into $\mathcal{O}$ with priority $\hat{s}$ }
}

\BlankLine
\While{$\mathcal{O}\neq\emptyset$ \textbf{and} $S<S_{\max}$}{
  extract node with maximal $\hat{s}$: $(p,d,\hat{s},h)$ from $\mathcal{O}$;\quad $S\leftarrow S+1$\;
  \If{$d \ge D_{\max}$ \textbf{or} $\max(\mathrm{height}(p),\mathrm{width}(p)) \le u_{\min}$ \textbf{or} $h \ge \tau_d$}{
    $\mathcal{P}_{\mathrm{sel}} \leftarrow \mathcal{P}_{\mathrm{sel}} \cup \{p\}$;\ \textbf{continue}}
  \tcp{expand into $3{\times}3$ children and score}
  divide $p$ into $\{p_k\}_{k=1}^9$;\quad $\mathcal{C}\!\leftarrow\!\emptyset$\;
  \For{$k=1$ \KwTo $9$}{ $(\hat{s}_k,h_k)\leftarrow$ ScorePatch$(p_k,q,d{+}1)$;\ 
    $\mathcal{C}\leftarrow \mathcal{C}\cup\{(p_k,d{+}1,\hat{s}_k,h_k)\}$ }
  sort $\mathcal{C}$ by $\hat{s}_k$ desc;\quad
  $\mathcal{C}_{\mathrm{keep}} \leftarrow \{\,\cdot\,\in\mathcal{C}\mid \hat{s}_k \ge \gamma\}$\;
  \If{$\mathcal{C}_{\mathrm{keep}}=\emptyset$}{ $\mathcal{C}_{\mathrm{keep}} \leftarrow$ top-1 of $\mathcal{C}$ }
  truncate $\mathcal{C}_{\mathrm{keep}}$ to at most $K_{\max}$ elements\;
  \ForEach{$(p_k,d{+}1,\hat{s}_k,h_k)\in\mathcal{C}_{\mathrm{keep}}$}{
    insert $(p_k,d{+}1,\hat{s}_k,h_k)$ into $\mathcal{O}$ with priority $\hat{s}_k$ }
}

\Return $\mathcal{P}_{\mathrm{sel}}$ \;
\end{algorithm*}

\subsection{Layout-Aware Patch Reassembly}
\label{app:algo:reassembly}

Given $\mathcal{P}_{\mathrm{sel}}$, we build a compact, layout-faithful canvas
$\tilde{I}$ in three stages, which is shown in Alg.~\ref{alg:lapr}:

\textbf{(1) Coarse partition.}
We assign selected patches to a coarse $3{\times}3$ partition
$\{\Omega_{i,j}\}_{i,j=0}^2$ and build a binary mask $M_{i,j}$ for each region, marking
occupied fine-grid cells. This anchors every patch to a stable quadrant
(top-left $\rightarrow$ bottom-right) preserved in the final layout.

\textbf{(2) Intra-region compression.}
For each $(i,j)$, we first locate all non-empty rows and columns in $M_{i,j}$.
Let $R_{i,j}$ be the indices of non-zero rows and $C_{i,j}$ the indices of non-zero columns.
We then form a compact mask $\hat{M}_{i,j}=M_{i,j}[R_{i,j},\,C_{i,j}]$ that drops only all-zero rows/columns.
Using $\hat{M}_{i,j}$ to index the original image $I$, we crop a tight content block $I_{i,j}$ with size $(h_{i,j},w_{i,j})$.
No reordering or resampling is applied at this stage—row/column order is preserved—so the relative placement of selected cells inside the region remains unchanged.
This preserves \emph{intra-region} spatial relations while removing purely empty margins.

\textbf{(3) Global rearrangement and padding.}
We set $H_c=\max_{i,j}h_{i,j}$ and $W_c=\max_{i,j}w_{i,j}$ and allocate a blank
$3H_c{\times}3W_c$ canvas $\tilde{I}$ (gray background). Each non-empty block $I_{i,j}$
is resized with aspect-ratio–preserving padding to $(H_c,W_c)$ and placed into its fixed
slot $\tilde{I}[iH_c:(i{+}1)H_c,\,jW_c:(j{+}1)W_c]$.
Thus \emph{global} quadrant location and \emph{local} arrangement are jointly preserved,
while the overall token footprint is tightly bounded.

\subsection{Complexity Analysis}
Let the long side of the input be $L_0$ and the minimum search unit be $u_{\min}$. Each zoom step shrinks the long side by $1/3$, so the depth to reach $u_{\min}$ along one path is
\begin{equation}
    D \;=\; \left\lceil \log_3\!\bigl(L_0/u_{\min}\bigr) \right\rceil .
\end{equation}
For a $20$K image with $u_{\min}{=}224$, this gives $D{=}5$. With adaptive branching, the total number of evaluated nodes is bounded by
$\min\!\bigl(S_{\max}, \sum_{d=0}^{D-1} K_{\max}^{\,d}\bigr)$, i.e., $O\!\bigl(\min(S_{\max}, (K_{\max}^{D}-1)/(K_{\max}-1))\bigr)$.

We bound the traversal by $S_{\max}$ nodes. Per-node complexity is driven by $\mathcal{R}$, with a low-overhead pruned-view query to $\mathcal{F}$, rendering the search practical for Ultra-HR inputs. Reassembly is linear in the number of
selected cells and involves only simple indexing and padding. In practice, we cache
tile-level features for $\mathcal{R}$, vectorize pruning and masking, and use a
priority-queue implementation for $\mathcal{O}$.

\begin{algorithm}[ht]
\caption{Layout-Aware Patch Reassembly}
\label{alg:lapr}
\small
\DontPrintSemicolon

\KwIn{image $I$; selected patch set $\mathcal{P}_{\mathrm{sel}}$ on the search grid; coarse $3{\times}3$ partition $\{\Omega_{i,j}\}_{i,j\in\{0,1,2\}}$}
\KwOut{reassembled canvas $\tilde{I}$}

\BlankLine
\For{$i,j \in \{0,1,2\}$}{
    $M_{i,j} \leftarrow \textsc{MaskFromPatches}\bigl(\mathcal{P}_{\mathrm{sel}} \cap \Omega_{i,j}\bigr)$\;
    $R_{i,j} \leftarrow \{\,r \mid \exists c:\; M_{i,j}(r,c)=1\,\}$\;
    $C_{i,j} \leftarrow \{\,c \mid \exists r:\; M_{i,j}(r,c)=1\,\}$\;

    \eIf{$R_{i,j}=\emptyset$ \textbf{or} $C_{i,j}=\emptyset$}{
        $I_{i,j} \leftarrow \varnothing$;\quad $(h_{i,j},w_{i,j}) \leftarrow (0,0)$\;
    }{
        \tcp{drop only all-zero rows/columns to preserve intra-region order}
        construct $\hat{M}_{i,j}\in\{0,1\}^{|R_{i,j}|\times|C_{i,j}|}$ by
        $\hat{M}_{i,j}(\tilde r,\tilde c)=M_{i,j}\!\bigl(R_{i,j}[\tilde r],\,C_{i,j}[\tilde c]\bigr)$\;
        $I_{i,j} \leftarrow \textsc{Crop}\!\bigl(I,\hat{M}_{i,j},\Omega_{i,j}\bigr)$\;
        $(h_{i,j},w_{i,j}) \leftarrow \textsc{Size}(I_{i,j})$\;
    }
}

$H_c \leftarrow \max_{i,j} h_{i,j}$;\quad $W_c \leftarrow \max_{i,j} w_{i,j}$\;
initialize $\tilde{I}$ of size $(3H_c,\,3W_c)$ with neutral gray pixels\;

\For{$i,j \in \{0,1,2\}$}{
    \If{$h_{i,j} > 0$}{
        $\tilde{I}_{i,j} \leftarrow \textsc{ResizePad}(I_{i,j},\,H_c,\,W_c)$\;
        place $\tilde{I}_{i,j}$ into
        $\tilde{I}[\,iH_c:(i{+}1)H_c,\; jW_c:(j{+}1)W_c\,]$\;
    }
}

\Return $\tilde{I}$\;

\end{algorithm}

\begin{table*}[!t]
\centering
\caption{Ablations on the Two Core Components.}
\label{tab:Two Core Components.}
\resizebox{\textwidth}{!}{%
\begin{tabular}{lccccccccc}
\toprule
\textbf{Method} &
\textbf{Rural/urban} &
\textbf{Count} &
\textbf{Reasoning} &
\textbf{Status} &
\textbf{Category} &
\textbf{Shape} &
\textbf{Color} &
\textbf{Background} &
\textbf{Avg.} \\
\midrule
LLaVA-ov-7b   & 49.71\% & 8.90\% & 19.85\% & 13.60\% & 20.74\% & 29.05\% & 44.23\% & 17.65\% & 25.47\% \\
+ Adaptive Multi-Branch Zoom Search & 48.54\% & \textbf{\textcolor{darkyellow}{12.23\%}} & 24.63\% & \textbf{\textcolor{darkgreen}{16.91\%}} & 19.63\% & 29.88\% & \textbf{\textcolor{darkyellow}{48.08\%}} & \textbf{\textcolor{darkgreen}{29.41\%}} & 28.66\%\\
+ Layout-Aware Patch Reassembly & \textbf{\textcolor{darkyellow}{50.29\%}} & 6.42\% & \textbf{\textcolor{darkyellow}{25.37\%}} & 15.81\% & \textbf{\textcolor{darkyellow}{22.22\%}} & \textbf{\textcolor{darkyellow}{32.78\%}} & \textbf{\textcolor{darkgreen}{53.83\%}} & \textbf{\textcolor{darkyellow}{25.00\%}} & \textbf{\textcolor{darkyellow}{28.97\%}}\\
+ both of the Above & \textbf{\textcolor{darkgreen}{59.06\%}} & \textbf{\textcolor{darkgreen}{15.95\%}} & \textbf{\textcolor{darkgreen}{27.21\%}} & \textbf{\textcolor{darkyellow}{16.54\%}} & \textbf{\textcolor{darkgreen}{22.59\%}} & \textbf{\textcolor{darkgreen}{41.08\%}} & 47.12\% & 23.53\% & \textbf{\textcolor{darkgreen}{31.64\%}} \\
\bottomrule
\end{tabular}%
}
\end{table*}

\section{Extended Experiments}
\label{app:ablations}
We provide additional experiments to further validate the design and generality of \textbf{ZoomSearch}.
Specifically, we (i) ablate the two core components—\textbf{Adaptive Multi-Branch Zoom Search} and \textbf{Layout-Aware Patch Reassembly};
(ii) assess plug-and-play compatibility with a variety of base models; and
(iii) examine robustness to different scoring models.
Unless otherwise noted, all experiments are conducted on the same 2,000-sample subset of LRS-VQA used in the pilot study, with \texttt{llava-onevision-qwen2-7B} as the default backbone.
\subsection{Ablations on the Two Core Components}
\label{app:abl_core}
To evaluate the roles of our two principal design components, we examine each in isolation before assessing their combined effect.
As reported in Table~\ref{tab:Two Core Components.}, using \textbf{Adaptive Multi-Branch Zoom Search} on its own leads to an average improvement of \textbf{12.52\%}. In this setting, the search mechanism is preserved but the selected crops are simply arranged sequentially without preserving the spatial layout.
We then isolate \textbf{Layout-Aware Patch Reassembly} by selecting patches solely according to patch–text relevance and subsequently reconstructing them with the layout-aware strategy. This configuration yields a slightly larger gain of \textbf{13.74\%}.
When both components are activated together, the average accuracy rises to \textbf{24.22\%}. This joint improvement demonstrates that the hierarchical search procedure and the spatially faithful reassembly mechanism contribute complementary strengths, and that their integration is essential to realizing the full capability of \textbf{ZoomSearch}.

\subsection{Plug-and-Play Across Different Base Models}
\label{app:abl_bases}

We evaluate \textbf{ZoomSearch} as a plug-and-play front end across diverse backbones—two remote sensing foundation models (GeoChat, VHM) and several general-purpose MLLMs from the LLaVA and Qwen-VL families. We report the results in Tab.~\ref{tab:plug_and_play}. Across all backbones, \textbf{ZoomSearch} improves the average accuracy: for example, gains of +\textbf{6.51\%} for GeoChat, +\textbf{7.15\%} for VHM, +\textbf{7.14\%} for Qwen2.5-VL-7b, +\textbf{12.12\%} for LLaVA-v1.6-7b, and +\textbf{24.22\%} for LLaVA-ov-7b are observed. The improvements are particularly pronounced on fine-grained categories (\emph{Category}, \emph{Shape}, \emph{Color}), suggesting that concentrating tokens on query-relevant regions is especially beneficial for detail-oriented reasoning.

The magnitude of improvement correlates with backbone quality. Stronger models (e.g., LLaVA-ov-7b, LLaVA-v1.6-7b) provide more reliable model-evidence signals for early stopping and branching, and thus realize larger overall gains. In contrast, LLaVA-v1.5-7b exhibits higher variance in the evidence signal and struggles to answer correctly even when the search localizes the right area, leading to a modest net gain. For remote sensing backbones, we observe solid average improvements despite a mild decline on \emph{Rural/urban}; this likely reflects a distribution shift introduced by our compact canvas relative to the models’ pretraining imagery for scene-level classification. Excluding this scene-type artifact, \textbf{ZoomSearch} consistently enhances object- and attribute-centric tasks, underscoring the method’s effectiveness and robustness across architectures and domains.

\begin{table*}[!t]
\centering
\caption{Plug-and-Play Across Different Base Models.}
\label{tab:plug_and_play}
\resizebox{\textwidth}{!}{%
\begin{tabular}{lccccccccc}
\toprule
\textbf{Method} &
\textbf{Rural/urban} &
\textbf{Count} &
\textbf{Reasoning} &
\textbf{Status} &
\textbf{Category} &
\textbf{Shape} &
\textbf{Color} &
\textbf{Background} &
\textbf{Avg.} \\
\midrule
GeoChat~\cite{geochat} & 65.50\% & 10.40\% & 16.91\% & 7.72\% & 5.93\% & 23.24\% & 18.75\% & 16.18\% & 20.58\%\\
\textit{+ ZoomSearch} & 66.08\% & 11.01\% & 17.65\% & 8.82\% & 8.52\% & 24.48\% & 19.71\% & 19.12\% & 21.92\% \\
\textit{Improvement} & \textbf{\textcolor{darkred}{+0.89\%}} & \textbf{\textcolor{darkred}{+5.87\%}} & \textbf{\textcolor{darkred}{+4.38\%}} & \textbf{\textcolor{darkred}{+14.25\%}} & \textbf{\textcolor{darkred}{+43.68\%}} & \textbf{\textcolor{darkred}{+5.34\%}} & \textbf{\textcolor{darkred}{+5.12\%}} & \textbf{\textcolor{darkred}{+18.17\%}} & \textbf{\textcolor{darkred}{+6.51\%}} \\
\addlinespace[2pt]
\hdashline                         
\addlinespace[2pt]
VHM~\cite{vhm} & 59.65\% & 10.70\% & 21.32\% & 14.34\% & 13.70\% & 32.37\% & 48.56\% & 17.65\% & 27.29\%\\
\textit{+ ZoomSearch} & 54.97\% & 11.31\% & 19.12\% & 18.01\% & 18.15\% & 39.42\% & 55.29\% & 17.65\% & 29.24\% \\
\textit{Improvement} & \textbf{\textcolor{darkgrey}{-7.85\%}} & \textbf{\textcolor{darkred}{+5.70\%}} & \textbf{\textcolor{darkgrey}{-10.32\%}} & \textbf{\textcolor{darkred}{+25.59\%}} & \textbf{\textcolor{darkred}{+32.48\%}} & \textbf{\textcolor{darkred}{+21.78\%}} & \textbf{\textcolor{darkred}{+13.86\%}} & \textbf{\textcolor{darkred}{0.00\%}} & \textbf{\textcolor{darkred}{+7.15\%}} \\
\addlinespace[2pt]
\hdashline                         
\addlinespace[2pt]
Qwen2.5-VL-7b~\cite{qwen2.5}  & 48.54\% & 13.46\% & 20.22\% & 8.82\% & 15.93\% & 18.67\% & 54.81\% & 23.53\% & 25.62\%\\
\textit{+ ZoomSearch} & 50.29\% & 11.96\% & 20.59\% & 9.56\% & 15.93\% & 32.37\% & 55.39\% & 23.53\% &27.45\%\\
\textit{Improvement} & \textbf{\textcolor{darkred}{+3.61\%}} & \textbf{\textcolor{darkgrey}{-11.14\%}} & \textbf{\textcolor{darkred}{+1.83\%}} & \textbf{\textcolor{darkred}{+8.39\%}} & \textbf{\textcolor{darkred}{0.00\%}} & \textbf{\textcolor{darkred}{+73.38\%}} & \textbf{\textcolor{darkred}{+1.06\%}} & \textbf{\textcolor{darkred}{0.00\%}} & \textbf{\textcolor{darkred}{+7.14\%}} \\
\addlinespace[2pt]
\hdashline                         
\addlinespace[2pt]
LLaVA-v1.6-7b~\cite{llavanext}  & 47.36\% & 7.95\% & 20.96\% & 18.38\% & 13.70\% & 31.95\% & 41.83\% & 20.59\% & 25.34\%\\
\textit{+ ZoomSearch}  & 50.58\% & 14.42\% & 20.96\% & 14.71\% & 15.19\% & 40.25\% & 44.71\% & 26.47\% & 28.41\%\\
\textit{Improvement} & \textbf{\textcolor{darkred}{+6.80\%}} & \textbf{\textcolor{darkred}{+81.38\%}} & \textbf{\textcolor{darkred}{0.00\%}} & \textbf{\textcolor{darkgrey}{-19.97\%}} & \textbf{\textcolor{darkred}{+10.88\%}} & \textbf{\textcolor{darkred}{+25.98\%}} & \textbf{\textcolor{darkred}{+6.89\%}} & \textbf{\textcolor{darkred}{+28.56\%}} & \textbf{\textcolor{darkred}{+12.12\%}} \\
\addlinespace[2pt]
\hdashline                         
\addlinespace[2pt]
LLaVA-ov-7b~\cite{llavaov}  & 49.71\% & 8.90\% & 19.85\% & 13.60\% & 20.74\% & 29.05\% & 44.23\% & 17.65\% & 25.47\% \\
\textit{+ ZoomSearch} & 59.06\% & 15.95\% & 27.21\% & 16.54\% & 22.59\% & 41.08\% & 47.12\% & 23.53\% & 31.64\% \\
\textit{Improvement} & \textbf{\textcolor{darkred}{+18.81\%}} & \textbf{\textcolor{darkred}{+79.21\%}} & \textbf{\textcolor{darkred}{+37.08\%}} & \textbf{\textcolor{darkred}{+21.62\%}} & \textbf{\textcolor{darkred}{+8.92\%}} & \textbf{\textcolor{darkred}{+41.41\%}} & \textbf{\textcolor{darkred}{+6.53\%}} & \textbf{\textcolor{darkred}{+33.31\%}} & \textbf{\textcolor{darkred}{+24.22\%}} \\
\bottomrule
\end{tabular}%
}
\end{table*}

\begin{table*}[!t]
\centering
\caption{Ablations on scoring models.}
\label{tab:scoring models}
\small
\resizebox{\textwidth}{!}{%
\begin{tabular}{lcccccccccc}
\toprule
\textbf{Method} &
\textbf{Rural/urban} &
\textbf{Count} &
\textbf{Reasoning} &
\textbf{Status} &
\textbf{Category} &
\textbf{Shape} &
\textbf{Color} &
\textbf{Background} &
\textbf{Avg.} \\
\midrule
LLaVA-ov-7b  & 49.71\% & 8.90\% & 19.85\% & \textbf{\textcolor{darkyellow}{13.60\%}} & 20.74\% & 29.05\% & 44.23\% & 17.65\% & 25.47\% \\
\midrule
+ RemoteCLIP~\cite{remoteclip} & \textbf{\textcolor{darkyellow}{60.52\%}} & 12.23\% & \textbf{\textcolor{darkyellow}{26.83\%}} & 12.13\% & \textbf{\textcolor{darkgreen}{22.96\%}} & 40.25\% & \textbf{\textcolor{darkgreen}{49.01\%}} & \textbf{\textcolor{darkgreen}{27.94\%}} & \textbf{\textcolor{darkyellow}{31.48\%}} \\
+ GeoRSCLIP~\cite{GeoRSCLIP} & 60.23\% & \textbf{\textcolor{darkyellow}{12.27\%}} & 20.96\% & 11.76\% & \textbf{\textcolor{darkyellow}{22.59\%}} & \textbf{\textcolor{darkyellow}{41.08\%}} & 47.12\% & 25.00\% & 30.13\% \\
+ LRSCLIP~\cite{lrsclip} & \textbf{\textcolor{darkgreen}{60.82\%}} & 11.31\% & \textbf{\textcolor{darkyellow}{23.90\%}} & \textbf{\textcolor{darkyellow}{13.60\%}} & 22.22\% & \textbf{\textcolor{darkgreen}{42.32\%}} & \textbf{\textcolor{darkyellow}{47.60\%}} & \textbf{\textcolor{darkyellow}{26.47\%}}  & 31.03\%\\
+ VisRAG~\cite{visrag} & 59.06\% & \textbf{\textcolor{darkgreen}{15.95\%}} & \textbf{\textcolor{darkgreen}{27.21\%}} & \textbf{\textcolor{darkgreen}{16.54\%}} & \textbf{\textcolor{darkyellow}{22.59\%}} & \textbf{\textcolor{darkyellow}{41.08\%}} & 47.12\% & 23.53\% & \textbf{\textcolor{darkgreen}{31.64\%}}  \\
\bottomrule
\end{tabular}%
}
\end{table*}

\subsection{Ablation on Scoring Models}
\label{subsec:ablation_scoring_models}
To compute the patch-text relevance term $g(t)$, we replace the scoring model $\mathcal{R}$ with four alternatives—RemoteCLIP~\cite{remoteclip}, GeoRSCLIP~\cite{GeoRSCLIP}, LRSCLIP~\cite{lrsclip}, and VisRAG~\cite{visrag}—while keeping the search/reassembly pipeline fixed. As shown in Table~\ref{tab:scoring models}, all variants yield consistent improvements over the backbone (\(\approx 22\%\)). Overall, VisRAG attains the highest average accuracy, but the margins are modest, suggesting that \textbf{ZoomSearch} is robust to the specific choice of $\mathcal{R}$. Unless otherwise noted, we therefore adopt VisRAG for main results.

\section{Additional Results}
\label{app:more_results}

\subsection{Additional Qualitative Examples}
\label{app:more_qual}

\begin{figure*}[!t]
    \centering
    \includegraphics[width=\textwidth]{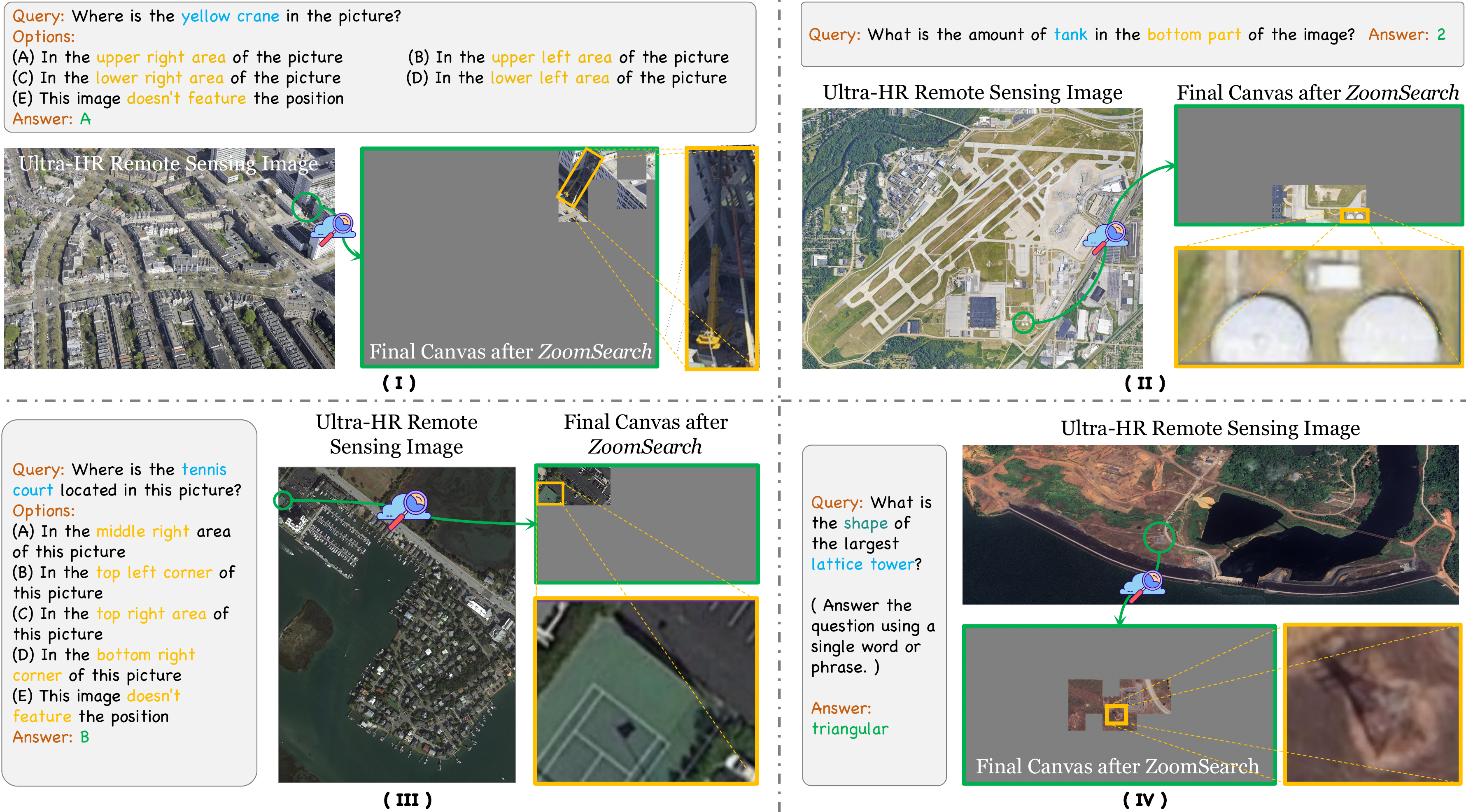}
    \caption{Qualitative results of \textbf{ZoomSearch}. In both examples, the method successfully zooms into extremely fine-grained target objects in Ultra-HR remote sensing images and produces correct answers.}
    \label{fig:good_cases}
\end{figure*}

Figure~\ref{fig:good_cases} presents four representative Ultra\mbox{-}HR RS\mbox{-}VQA cases. 
\textbf{(I)} and \textbf{(III)} depict dense urban scenes with minute targets. \textbf{ZoomSearch} first localizes the \emph{yellow crane} and the \emph{tennis court}, respectively; layout-aware reassembly then anchors each region to its original quadrant, preserving \emph{global} placement. This preservation of absolute orientation is crucial for answering \emph{directional} queries (e.g., ``upper right area,'' ``to the right above the picture''), which are prevalent in remote sensing tasks that require image-referenced positioning, including incident reporting, infrastructure inspection, and asset monitoring.
\textbf{(II)} considers a counting scenario, in which \textbf{ZoomSearch} successfully localizes all \emph{tanks}. By preserving each patch’s absolute position in the original frame, the reassembled canvas renders spatial descriptors such as ``bottom part of the image'' unambiguous during answering.
\textbf{(IV)} involves a low\mbox{-}contrast \emph{lattice tower} amid clutter. Hierarchical zooming concentrates evidence on the subtle structure, and the final layout\mbox{-}faithful canvas supports a precise prediction.

\begin{figure*}[!t]
    \centering
    \includegraphics[width=\textwidth]{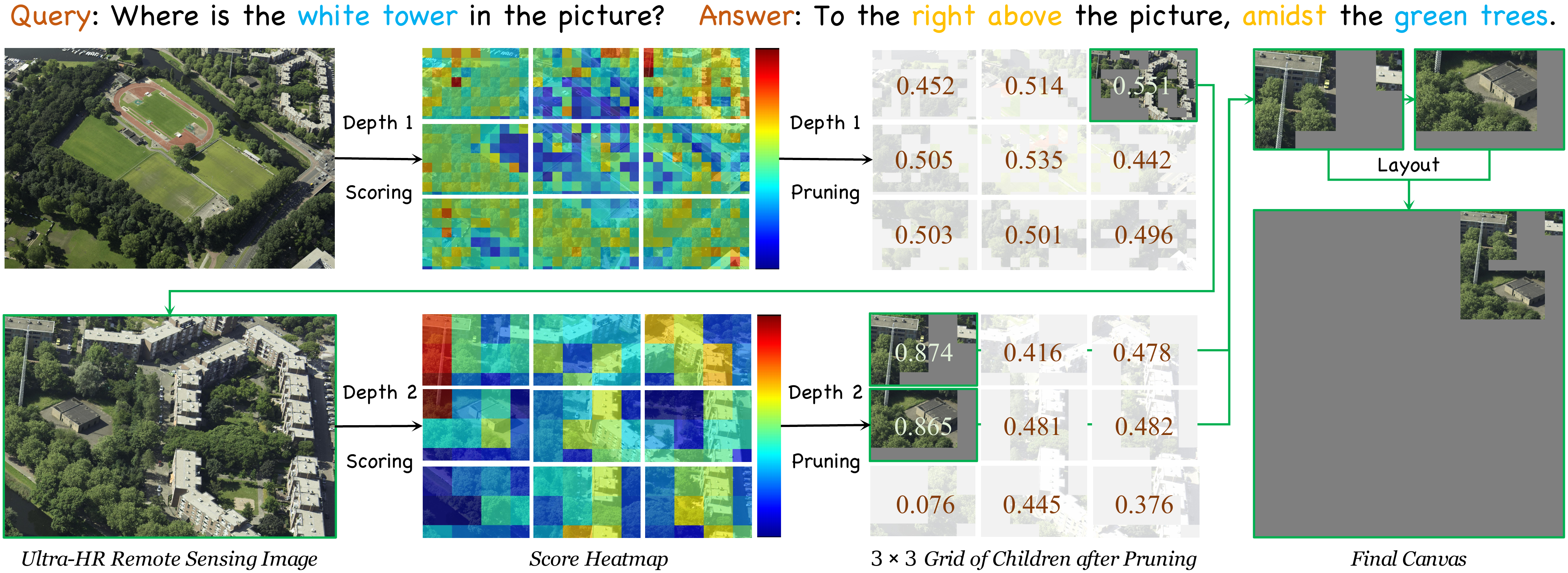}
    \caption{\textbf{Overview of the ZoomSearch pipeline.} Given an Ultra-HR remote sensing image and a query, the procedure (i) scores a $3{\times}3$ grid and prunes 50\% low-confidence tiles, (ii) performs hierarchical expansion with adaptive Top-$k$ retention, and (iii) reassembles the selected patches into a compact, layout-faithful canvas that preserves both global orientation and intra-region spatial relations. The example illustrates localization of the \emph{white tower} and construction of the final canvas.}
    \label{fig:pipeline}
\end{figure*}

Figure~\ref{fig:pipeline} illustrates the \textbf{ZoomSearch} pipeline for a query about the \emph{white tower}’s location. Depth-1 heatmaps already highlight the correct neighborhood, though some tiles receive spurious high scores. As the search continues, hierarchical zooming and progressive pruning suppress these false positives and strengthen evidence around the true region. Although the object spans two tiles at Depth 2, the \emph{adaptive Top-$k$} policy keeps both branches, ensuring full coverage. The final layout-aware canvas preserves global orientation and local spatial relations, enabling an accurate answer.

\subsection{Failure Cases and Analysis}
\label{app:failures}

\begin{figure*}[!t]
    \centering
    \includegraphics[width=\textwidth]{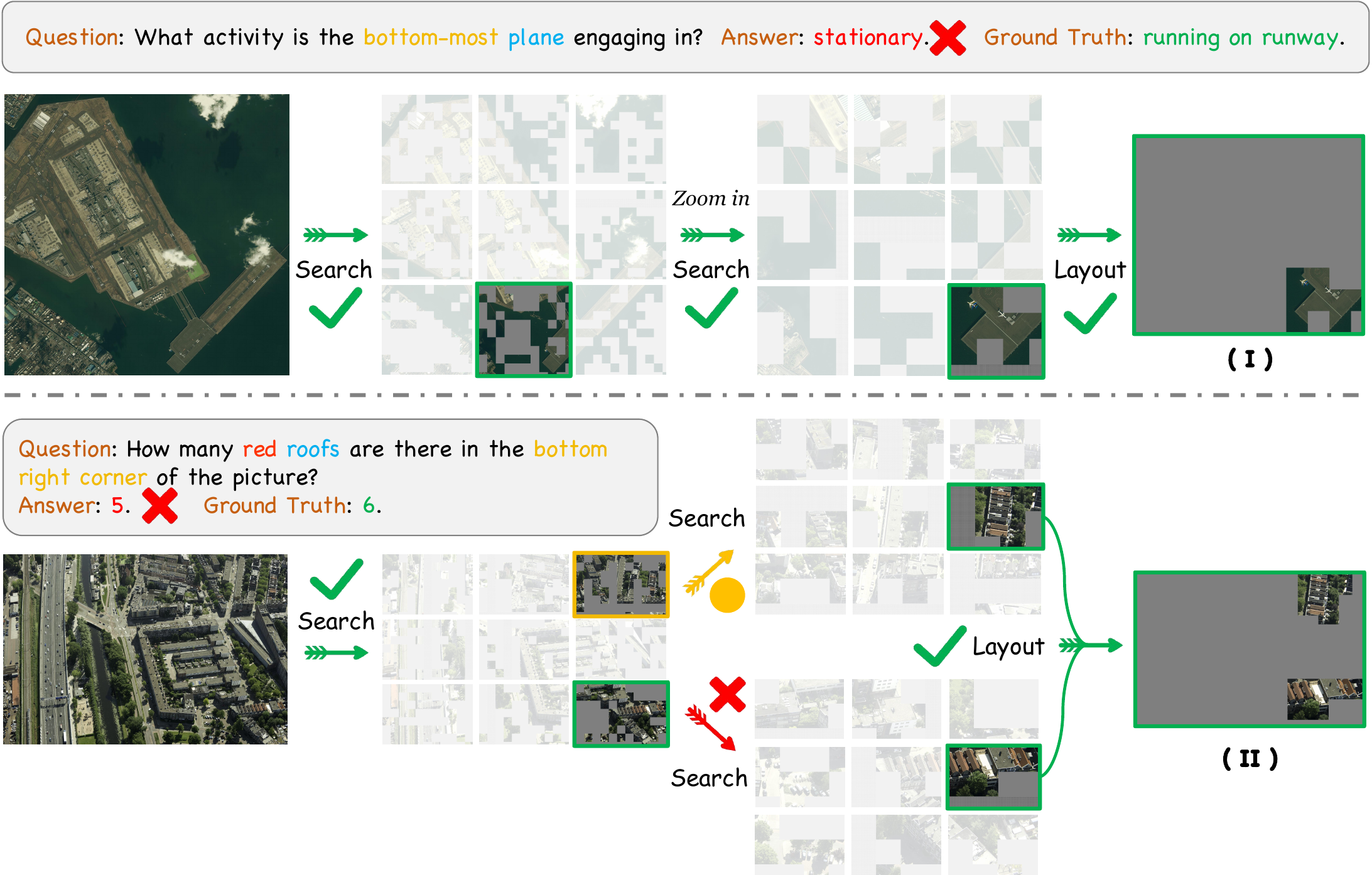}
    \caption{Failure cases of \textbf{ZoomSearch}. \textbf{(I)} The search correctly localizes and lays out the target region, yet the backbone produces an incorrect answer. \textbf{(II)} Incomplete search: a small red roof at the far left is not retained, yielding an undercount and a wrong prediction.}
    \label{fig:failure_cases}
\end{figure*}

Figure~\ref{fig:failure_cases} highlights two recurrent failure modes observed in our evaluation.
\textbf{(I) Backbone–limited answering.} The search correctly isolates the target region, and the reassembly preserves both its global orientation and intra-region spatial relations, yet the base VLM still produces an incorrect answer. In such cases the visual evidence is present and well framed, but the foundation model’s recognition or reasoning is insufficient. This suggests an upper bound imposed by the backbone rather than the search policy. \textbf{(II) Incomplete or imperfect search.} The procedure occasionally misses a small, relevant instance (\emph{under-coverage}) or retains a distractor (\emph{over-coverage}). The example shows one red roof on the far left not selected at Depth 3, leading to an undercount. Notably, the layout anchors patches to their original quadrants. Consequently, the instruction ‘bottom right corner’ constrains the backbone to count only within the bottom-right canvas block. This constraint mitigates the error, although it does not eliminate it entirely.

\textbf{Future work} may explore (i) more robust search policies that better balance coverage and precision under Ultra-HR remote sensing complexity, (ii) lightweight self-verification modules that assess whether the retrieved patches adequately cover the query-relevant evidence—thereby reducing error accumulation from imperfect search, and (iii) learned scoring mechanisms that couple retrieval signals more tightly with the backbone’s internal representations.

{
    \small
    \bibliographystyle{ieeenat_fullname}
    \bibliography{main}
}

\end{document}